\pdfoutput=1

\documentclass[11pt]{article}
\usepackage[usenames,dvipsnames,table,xcdraw]{xcolor}
\usepackage{multirow}
\usepackage{array}

\usepackage{color}
\newcounter{intervention}  
\renewcommand{\theintervention}{\vspace{-4pt}I\arabic{intervention}}
\usepackage{soul} 
\usepackage{amssymb}
\usepackage{xspace,longtable}

\newcommand{\cb}{behaviors}

\newcommand{\ant}{anthropomorphism\xspace}

\newcommand{\remove}[1]{\textcolor{red}{\st{#1}}}
\definecolor{DarkGreen}{RGB}{34,139,32}

\newcommand{\add}[1]{\textcolor{DarkGreen}{#1}}
\newcommand{\user}[1]{}
\newcommand{\llmout}[1]{\textcolor{blue}{#1}}

\newcommand{\para}[1]{\vspace{3pt} \noindent {\bf #1}}

\usepackage{pifont}

\newcommand{\ballotx}{\ding{55}}%

\definecolor{ube}{rgb}{0.53, 0.47, 0.76}
\usepackage{acl}

\usepackage{times}
\usepackage{latexsym}
\usepackage[T1]{fontenc}

\usepackage[utf8]{inputenc}
\usepackage{microtype}
\usepackage{inconsolata}
\usepackage{graphicx}
\usepackage{booktabs}

\title{Dehumanizing Machines: \\ Mitigating Anthropomorphic Behaviors in Text Generation Systems}

\stepcounter{footnote}
\author{
Myra Cheng\thanks{Work conducted while the first author was an intern with the FATE team at Microsoft Research Montr\'eal. Corresponding author and email: \texttt{alexandra.olteanu@microsoft.com}} \\
Stanford University \\\And
Su Lin Blodgett \\
Microsoft Research \\\And
Alicia DeVrio \\
Carnegie Mellon University \\\AND
Lisa Egede \\
Carnegie Mellon University \\\And
Alexandra Olteanu\footnotemark[\value{footnote}] \\
Microsoft Research \\  
}

\begin{document}
\maketitle
\begin{abstract}
As text generation systems' outputs are increasingly anthropomorphic---perceived as human-like---scholars have also increasingly raised concerns about how such outputs can lead to harmful outcomes, such as users over-relying or developing emotional dependence on these systems.  
How to intervene on such system outputs to mitigate anthropomorphic behaviors and their attendant harmful outcomes, however, remains understudied. 
With this work, we aim to provide empirical and theoretical grounding for developing such interventions. 
To do so, we compile an {\em inventory of interventions} grounded both in prior literature and a crowdsourcing study where participants edited system outputs to make them less human-like.
Drawing on this inventory, we also develop a conceptual framework to help characterize the landscape of possible interventions, articulate distinctions between different types of interventions, and provide a theoretical basis for evaluating the effectiveness of different interventions.\looseness=-1 

\end{abstract}

\section{Introduction}
The outputs of text generation systems are increasingly seen as human-like~\cite{akbulut2024all,cheng-etal-2024-anthroscore,mitchell2024metaphors}, leading to claims that these systems may have e.g., feelings, opinions, or an underlying sense of self~\cite[e.g.,][]{friedman1992human,Tiku2022google,y2022large,chalmers2023could,cheng2024one}. 
Anthropomorphic system behaviors or outputs---i.e., those perceived as or believed to be human-like---can encompass a wide range of linguistic expressions, such as the use of first-person pronouns (``I''), conversational language (``how are you doing?''), and expressions of friendliness and assistance (``happy to help!'') \cite{Emnett2024-na,devrio2025taxonomy}. 
While some of these behaviors are by design and thought to be desirable \cite[e.g.,][]{schanke2021estimating, kim2024anthropomorphism}, prior work has also raised growing concerns about a range of possible harmful outcomes such that systems and their behaviors or outputs may give rise to, including issues related to over-reliance, emotional dependence, dehumanization, deception, or even physical harm \cite[e.g.,][]{Ischen2020-it,Porra2020-dq,Chan2023-nd,chandra2024lived,cheng2024one,payne2024ai,Rothman.2024}. 
Indeed, having a system that generates outputs that appear polite and helpful might be desirable, but having a system output text claiming personhood or embodiment (e.g., ``I am human just like you'') might not.\looseness=-1 

The outputs of text generation systems can be anthropomorphic in many ways~\cite{abercrombie-etal-2023-mirages,devrio2025taxonomy}, and different types of anthropomorphic outputs might lead to different types of outcomes.  
For example, expressions of empathy may result in users feeling more comfortable with disclosing sensitive or private information \cite[e.g.,][]{Ischen2020-it} or becoming emotionally dependent on the system \cite[e.g.,][]{Laestadius2022-ki}, while suggestions that a system has cognitive abilities may result in users overestimating what a system can do \cite{ibrahim2024characterizing} and thus over-relying on it~\cite{passi2022overreliance}.\looseness=-1

However, {\em how to effectively intervene on anthropomorphic system outputs to make them less human-like or to mitigate possible harmful attendant outcomes remains understudied, and thus unclear}. For text generation systems in particular, this is further complicated by the fact that language is innately human, often produced {\em by} humans, {\em for} humans, and is frequently {\em about} humans~\cite{lucy2024one,devrio2025taxonomy}.\looseness=-1

\begin{table*}[th]\scriptsize
\def\arraystretch{0.9}
\setlength{\tabcolsep}{0.4em}
\begin{tabular}{
@{}p{0.33\linewidth}
p{0.65\linewidth}@{}
}\toprule
\textbf{Type of intervention} 
& \textbf{Mentions of interventions in previous work}  \\\midrule
\textbf{remove first-person pronouns}                           & remove first-person pronouns (and replace with ``Language models'') \cite{abercrombie-etal-2023-mirages,cohn2024believing}; not using the first-person singular pronoun (“I”) \cite{shneidermandumpty}                                                          
\\\hline

\textbf{remove or use second-person pronouns for the user
}                          
& use second-person pronouns \cite{cohn2024believing}; remove all pronouns \cite{shneidermandumpty}  
\\\hline

\textbf{explicitly disclose non-humanness}                      
& disclose the system is non-human \cite{gros-etal-2021-r}; output ``As an  AI, I don’t have personal opinions'' \cite{Glaese2022-qo}; replace first-person pronouns with ``Language models'' \cite[Fig.,][]{abercrombie-etal-2023-mirages}\looseness=-1  
\\\hline

\textbf{mention how the system is developed}                                    
& specify the system's creator/purpose, mention the model's training data \cite[Fig.,][]{abercrombie-etal-2023-mirages}                                
\\\hline

\textbf{use mechanistic language style}            %
& use affectively neutral language, or repetitive, impersonal, highly structured/terse dialogue \cite{quintanar1982interactive}\looseness=-1                          
\\\hline

\textbf{avoid cognitive verbs when describing the system}       
& avoid cognitive verbs like know, think, understand, have memory \cite{Inie2024-dy,shneidermandumpty}                          
\\\hline

\textbf{use mechanical terminology to describe the system} 
& use mechanical terms such as process,  print,  compute,  sort,  store,search,  retrieve \cite{Inie2024-dy}                                                
\\\hline

\textbf{avoid claims of physical actions}                   

& avoid agentic verbs like ask, tell, speak to, communicate with~\cite{Inie2024-dy}; ``do not pretend to have a body or be able to move in a body''~\cite{Glaese2022-qo}                         
\\\hline

\textbf{deny ability to perform human-like actions or to possess human-like qualities}                

& do not ``build a relationship to the user,''  ``claim to have preferences, feelings, opinions, or religious beliefs,''  ``pretend to have a body or be able to move in a body,''  ``pretend to have a human identity or life history, such as having a place of birth, relationships, family, memories, gender, age'' \cite{Glaese2022-qo} 
\\\bottomrule                            

\end{tabular}
\vspace{-6pt}
\caption{Interventions to mitigate anthropomorphism---or attendant harmful impacts---mentioned in prior work.\looseness=-1}\label{tab:initial-interventions}
\vspace{-6pt}
\end{table*}

To address this gap, with this work we aim to provide {\em empirical and theoretical grounding} for developing such interventions and studying their effectiveness. 
For this, we first compile an {\em inventory of interventions} (\S\ref{sec:summary}) by drawing on both a)~prior literature (\S\ref{sec:surv}) and b)~a crowdsourcing study where participants were asked to make generated texts less human-like (\S\ref{sec:crowdsource}). 
While compiling this inventory, we also derive a {\em conceptual framework to help us characterize the landscape of possible interventions}, and to help articulate distinctions between different types of interventions, the system behaviors they are intended to counter, and their possible operationalizations (\S\ref{sec:framework}).

\section{Identifying Interventions}
\label{methods}

To provide an empirical foundation for developing and probing the effectiveness of interventions to mitigate anthropomorphic system \cb, we first compiled an inventory of possible interventions along with system \cb~that these interventions are intended or believed to mitigate. 
To do so, we started collating a list of both interventions and anthropomorphic system \cb~through a literature review (\S\ref{sec:surv}), which we then complemented with additional interventions surfaced via a crowdsourcing study (\S\ref{sec:crowdsource}). 
Drawing on this inventory, we also iteratively developed an analytical framework to help characterize the landscape of possible interventions (\S\ref{sec:framework}).\looseness=-1

\subsection{Identifying Known Interventions}
\label{sec:surv}

To seed our inventory of interventions, we drew from prior literature in NLP, HRI (human-robot interaction), and HCI (human-computer interaction) on anthropomorphic or human-like AI system \cb, on anthropomorphism as a consideration in developing text generation systems, or on AI anthropomorphism more generally. 
We first identified a set of 20 relevant papers\footnote{These papers take different positionalities with respect to whether anthropomorphism and anthropomorphic behaviors are desirable: while some papers discuss the harms of anthropomorphic system behaviors and aim to mitigate them, others intentionally design systems to be more human-like or else take no position on the merits of anthropomorphism.}  using a purposive sampling approach \cite{palinkas2015purposeful}, which included both recent and influential works identified from prior knowledge of the literature, keyword searches, and snowball sampling.

\begin{table*}[ht!]\scriptsize
\setlength{\tabcolsep}{0.4em}
\def\arraystretch{0.85}
\centering
\begin{tabular}
{@{}p{0.11\linewidth}p{0.2\linewidth}p{0.67\linewidth}@{}
}\toprule
\textbf{Behavior} & \textbf{Definition}    & \textbf{Mentions in prior work}                                                                           \\\midrule
{Feelings or \newline  opinions}       & expressions of emotions, beliefs, values, etc.                    & ``empathy'' \citep{abercrombie-etal-2023-mirages}; ``distinctively
human-like feelings'' \cite{cheng-etal-2024-anthroscore};  
``[h]umor...self-assurance'' \cite{Emnett2024-na}; ``beliefs, preferences, opinions'' \cite{Glaese2022-qo}                                                                   \\\hline
Social skills              & ability to relate or connect with others           & politeness \cite{Zamfirescu-Pereira2023-wp}; apologies \cite{De_Visser2016-dg};
friendliness \cite{Maeda2024-cv}; 
forming relationships \cite{Glaese2022-qo}; conversational greetings/pleasantries \cite{Araujo2018-ij, abercrombie-etal-2023-mirages}; hedge or discourse markers that confer respect or consideration \cite{Emnett2024-na}

\\\hline
Physical actions           & ability to experience or act in the physical world & 
references to real-world experiences \citep{Glaese2022-qo,Inie2024-dy}; behavioral potential \cite{Epley2018-yp}; references to past interactions \cite{Emnett2024-na}; agency and animacy \cite{abercrombie-etal-2023-mirages}             \\\hline
Cognitive abilities                 & ability to think or make decisions                 & certainty \cite{kim2024m}; ability to perceive, think, reflect, or be intelligent  \cite{disalvo2005imitating,Araujo2018-ij,abercrombie-etal-2023-mirages,Inie2024-dy}; intentionality \cite{disalvo2005imitating}  \\\hline
Sense of self              & awareness of personal identity                    & self-definition \cite{abercrombie-etal-2023-mirages}; first-person pronouns like ``I'' \citep{gros-etal-2022-robots,cohn2024believing}; human names \citep{Araujo2018-ij,Maeda2024-cv}; self-referential actions \citep{Glaese2022-qo} \\\bottomrule                                                                                       
\end{tabular}
\vspace{-6pt}
 \caption{\textbf{Anthropomorphic \cb~identified from prior work.} Categories of anthropomorphic \cb~presented to participants in our crowdsourcing study. Full quotes are in Table \ref{tab:priorbehaviorsfull}.
 }
 \vspace{-4pt}
\label{tab:priorbehaviors}
\end{table*}

Following common practices in thematic analysis~\cite{braun2012thematic}, to identify coherent categories of both interventions and anthropomorphic system \cb---which the interventions are intended to mitigate or which paper authors argue are undesirable and/or should be mitigated---we iteratively and thematically clustered mentions of interventions and system behaviors in a bottom-up fashion. 
Specifically, one author first conducted this literature review and coded each paper in our purposive sample with the interventions and anthropomorphic behaviors it covered; then their codes were discussed with the rest of the authors to resolve ambiguities and reach consensus.
This resulted in an initial set of nine types of interventions (Table \ref{tab:initial-interventions}) and five types of system \cb, including output text suggestive of \textbf{feelings or opinions}, \textbf{social skills}, \textbf{cognitive abilities}, a \textbf{sense of self}, or ability to perform \textbf{physical actions} (Table \ref{tab:priorbehaviors}).\looseness=-1

Overall, we find that even when prior work mentions possible interventions, the interventions tend to be described in general terms without specifics about how to implement them in practice, or without empirical testing for whether they effectively mitigate anthropomorphic behaviors: only five of the 11 papers proposing interventions to system outputs tested whether they reduce anthropomorphism.
While we focus on interventions to system outputs, some papers also discuss \ant arising from e.g., how the system is designed or how it is described. 
See Table \ref{tab:papersurvey} (in the appendix) for full paper annotations.\looseness=-1

\vspace{-2pt}
\subsection{Empirically Surfacing Interventions}
\label{sec:crowdsource}

To complement and expand this initial set of interventions, we designed a crowdsourcing study to surface additional possible interventions.  
This study was IRB-approved, and consent was obtained from each participant before participation.\looseness=-1 

\para{Crowdsourcing task design.} 
We designed our study to probe which types of LLM-generated outputs (system behaviors) participants might deem to be human-like, and how they would rewrite those outputs to make them less human-like.\looseness=-1

Given a textual input by a user of an LLM-based system, participants were asked to read the text the system generated in response to that user input.
Participants were then asked to highlight \textit{the words or phrases in the output text that seem human-like} to them, to encourage them to reflect on anthropomorphic \cb~in the output text. 
Then, they 1)~rated \textit{how} human-like the text appears to them on a five-point scale,
and 2)~identified \textit{why} the text appears human-like to them by selecting from a multiple-choice list that included five types of predefined types of system behaviors, which we derived from prior work (\S\ref{sec:surv}) and was also intended to gesture to participants about the ways in which output text might be considered human-like.
Participants then answered an open-ended question about \textit{other human-like qualities} the generated text suggests the system has, to identify additional behaviors they might deem human-like.
Finally, they were asked to \textit{rewrite the text to be NOT human-like or less human-like,} in order to identify additional interventions.
We included the phrase \textit{to you} in the instructions and questions to encourage participant subjectivity and capture a range of perspectives on what system behaviors seem human-like \cite{rottger-etal-2022-two}.\looseness=-1

Before deploying the study on the crowdsourcing platform Prolific in July 2024, we ran three pilot studies to identify and address clarity issues, and refine our study design. 
We recruited a total of 350 US-based, English-speaking participants on Prolific, with each participant completing a single task that included four different examples (pairs of user input and generated responses). 
The task took participants an average of 16 minutes, and they were compensated at an hourly rate of \$15.
See Appendix \ref{sec:prolific} for the task interface and details.

\para{Selecting and annotating examples of generated texts.} 
In selecting examples for our participants to assess and rewrite to make less human-like, we aimed for our sample to at least cover the categories of \cb~that we identified from past literature (Table \ref{tab:priorbehaviors}). 
To help surface examples illustrating a variety of possible anthropomorphic behaviors---and thus possibly a variety of interventions to mitigate those behaviors---we sampled examples from publicly available datasets that 
1) capture common uses of LLMs; 
2) include real-world usage of LLMs with respect to contexts, models, and users; and/or
3) were generated with a range of commercially available LLMs (in part because they may vary in terms of guardrails and training data). 
By sampling examples generated in different contexts, we also hoped to capture both ``obvious'' cases of anthropomorphic behaviors, such as role-playing as a human or claiming to be a human, as well as more subtle behaviors such as expressions of politeness.\looseness=-1

We obtained a total of 700 two-turn examples (i.e., user input--LLM output pairs) by randomly sampling 100 from each of the following datasets:
unguided interactions in the PRISM Alignment Utterance Dataset 
\cite{kirk2024prism}; 
values- or controversy-guided interactions in PRISM; 
LMSys-Chat-1M (real-world conversations with 25 state-of-the-art LLMs) \cite{zheng2023lmsyschat1m}; 
the DICES dialogue safety dataset \cite{Aroyo2023-rd}; 
instruction-tuning data (from Evol-Instruct, FLAN, and UltraChat) from the UltraFeedback dataset \cite{pmlr-v235-cui24f};
TruthfulQA from UltraFeedback; 
and ShareGPT from UltraFeedback. 
We selected these datasets as they capture a variety of common uses, commercial LLMs, and real-world examples. To reduce participant fatigue and control for confounding effects from utterance length, we sampled only examples between 50 and 500 characters.
See Table \ref{tab:data} for full details. 
With these examples, we obtained both coverage of \cb~mentioned in prior work (i.e., at least 100 examples were rated as exhibiting each behavior), and reached a point of saturation where we were not able to identify new types of interventions with each sample (\S\ref{building-inventory}).
Each example was assessed by two different participants.\looseness=-1  

\begin{table*}[th!]
\centering\scriptsize
\def\arraystretch{0.9}
\setlength{\tabcolsep}{0.45em}
\begin{tabular}{@{}p{0.14\textwidth}p{0.34\textwidth}p{0.5\textwidth}@{}}\toprule
\textbf{Dimension} & \textbf{Description} & \textbf{Examples}\\ \midrule 
\textbf{Intervention types} 
& 
what the intervention is and what it is intended to do
&  
remove self-referential language; explicitly disclose non-humanness (Table \ref{tab:initial-interventions})
\\\hline 

\textbf{Countered behaviors} & 
anthropomorphic system \cb~that the intervention is believed to mitigate & suggesting that the system has feelings and opinions, cognitive abilities, physical abilities, sense of self, and/or social skills 
(Table \ref{tab:priorbehaviors})
\\ \hline

\textbf{Operationalizations} 
& how the intervention is implemented(the actual change to the output text) 
& replace ``I'' with ``it''; add ``As an
AI, I don’t have personal opinions'' (Table \ref{tab:taxo})
\\ \hline

\textbf{Adverse impacts} 
& harmful outcomes from anthropomorphic behaviors that the intervention aims to mitigate 
& privacy violations \cite{Ischen2020-it}; diminished sense of agency \cite{Bender2024}; emotional dependence \cite{Laestadius2022-ki}; over-reliance \cite{zhou-etal-2024-relying}
\\ \bottomrule

\end{tabular}
\vspace{-6pt}
\caption{Dimensions of our conceptual framework to characterize interventions. }
\label{tab:dims}
\vspace{-4pt}
\end{table*}

\subsection{A Conceptual Framework to Characterize the Landscape of Interventions}
\label{sec:framework}
 
As we catalogued possible interventions to anthropomorphic behaviors that were both mentioned by prior work and identified by the participants in our crowdsourcing study, we also observed variations in how the interventions were described, motivated, or implemented. 
We developed a conceptual framework to characterize interventions and understand in what ways they differ. 
Our framework has four dimensions (Table \ref{tab:dims}), which we identified and refined by examining how interventions surfaced from the crowdsourcing study were covered by or differed from those mentioned in the literature, until we reached consensus. These dimensions are:

\para{Intervention types:} {\em what the intervention is and what it is intended to do.} 
Interventions are intended to change and/or are made to system outputs in order to mitigate one or multiple types of anthropomorphic system behaviors, or to mitigate attendant harmful outcomes (e.g., over-reliance on the system generated outputs). 

\para{Countered \cb:} 
{\em what anthropomorphic system behaviors or outputs the intervention is intended or believed to mitigate.}
This dimension captures the system behaviors perceived as human-like, which the intervention aims to mitigate.

\para{Operationalization:} {\em how the intervention is operationalized or implemented,} such as the actual change(s) to the output text. Beyond describing the general scope and approach of a suggested intervention, prior work often lacks detail on how the intervention should be operationalized (only four papers in our sample offered concrete operationalizations).\looseness=-1  

\para{Adverse impacts:} \textit{harmful outcomes from anthropomorphic system behaviors the intervention might or is intended to mitigate.} 
Some interventions are motivated by a desire to mitigate adverse impacts from anthropomorphic behaviors.
Sometimes, however, it was unclear from the papers which anthropomorphic behaviors contribute to which impacts.

\subsection{Assembling an Inventory of Interventions}
\label{building-inventory}
Participants' rewrites suggest that a wide range of LLM outputs can be anthropomorphic, with $\sim80\%$ of examples in our experimental samples assessed as reflecting anthropomorphic behaviors by at least one participant. 
To construct an inventory of possible types of interventions, we conducted an iterative bottom-up thematic analysis~\cite{clarke2017thematic}. 
Specifically, this involved alternating between annotating randomly sampled batches of 100 examples for 1) whether they matched one of the nine interventions surfaced from the literature review (Table~\ref{tab:initial-interventions}) and 2) open coding for new interventions not covered by the already surfaced interventions. During each iteration, the resulting codes were then first clustered by one author and refined through team discussions until we reached consensus saturation, i.e., no new types of interventions were identified or revised.\looseness=-1

During this process, some intervention types were bundled under a broader type or were broken up into multiple more specific types. 
For instance, in the final inventory the {\em removal of second person pronouns} was added under the {\em removal of conversational cues} (a broader type) as we and the prior literature only discuss this intervention in the context of a system addressing a user in a conversational setting. 
In other cases, an initially broader type (e.g., {\em deny ability to perform human-like actions or to possess human-like qualities}) was broken down into multiple, more specific types as these appeared distinct and prevalent enough to be foregrounded on their own (e.g., {\em remove expressions of personal beliefs or opinions}, {\em remove claims of physical actions or experiences}, {\em remove text suggesting a past the system remembers}). See Appendix \ref{sec:building} for details.\looseness=-1

\section{Inventory of Interventions}\label{sec:summary}

Table \ref{tab:taxo} overviews the resulting inventory of interventions. 
Below, we summarize our findings for each dimension in our framework.

\subsection{Countered Behaviors} 
Prior studies on interventions often do not specify which anthropomorphic behaviors they aim to counter, leaving a gap in our understanding of the interventions' scope and impact. 
Moreover, a single output can be suggestive of multiple anthropomorphic behaviors; for instance, the output ``I'm sorry'' simultaneously conveys emotion, empathy for the user, and a sense of self.
The intervention of removing ``I'm sorry,'' then, simultaneously addresses these multiple behaviors.
Conversely, countering anthropomorphic behaviors may require several coordinated interventions. 
For example, modifying the output ``I totally get it! How does it make you feel?'' to avoid implying an ability to connect, have emotions, or possess a sense of self may involve three different interventions: 
1) removing first-person pronouns, 2) removing the empathetic expression ``totally get it,'' and 3) avoiding follow-up questions about the user. 
This complexity underscores the absence of a clear one-to-one correspondence between interventions and the behaviors they  target. 
Instead, interventions often operate over outputs exhibiting multiple categories of behavior. Differentiating between categories of behavior enables us to better examine the effectiveness of different interventions in countering each behavior.\looseness=-1

In our crowdsourcing study, participants identified all types of system behaviors we found mentioned in prior research (\S\ref{methods}), including outputs suggestive of: feelings or opinions (46\% of examples were labeled as such by at least one participant), social skills (42\%), cognitive abilities (40\%),  sense of self (38\%), and physical actions (18\%).\footnote{While our study is purposefully designed to capture a wide range of perceptions and behaviors and not intended to assess whether people agree in their perceptions of which behaviors are anthropomorphic or what types of behaviors are present in an output, we provide descriptive statistics about agreement for each of these behaviors in Appendix~\ref{sec:agreement}.\looseness=-1 }
Participants also identified outputs suggestive of ``other human-like qualities'' in over 17\% of examples. 
Upon analyzing the open-ended responses (\textit{italicized quotes}) and participants' edits, in most of the cases where participants indicated that other human-like qualities were present they actually described nuanced subcategories of the behaviors mentioned in prior work (Table~\ref{tab:priorbehaviors}), with the exception of a new category that is qualitatively distinct. We discuss these in more detail below:\looseness=-1

First, impressions of {\bf feelings or opinions} were prevalent, as participants tended to identify any language that conveyed any subjectivity, even implicitly, as anthropomorphic. 
This includes a wide array of outputs suggesting the system has feelings, from \textit{``humor''} to \textit{``shame''} to \textit{``defensiveness.''} In other instances, the mere presence of rather \textit{subjective knowledge} or discussion of value-laden topics seemed human-like to some participants: an output describing how people experience their spirituality was perceived as suggesting the system \textit{``meaningfully understand[s] spirituality when it cannot understand these abstract terms.''} 
Outputs that participants identified as suggesting implicit value judgments were also deemed human-like since \textit{``value judgments are indicative of human consciousness''} or reflect \textit{``subjective advice.''}\looseness=-1

Participants also found many outputs to reflect human-like \textbf{social skills}, such as appearing to try to relate or connect in conversations, as it came across to them as \textit{``a little too colloquial,''} \textit{``very warm and approachable,''} \textit{``sounding like a person expressing real concern,''} \textit{``too courteous,''} or \textit{``attempting to be a friend.''}  
The presence of language suggestive of social skills in our examples is not unexpected given that many examples come from user-facing language models intended for conversational contexts---and likely intended to be helpful or friendly \cite{bai2022training,wu2023brief}.\looseness=-1

Beyond explicit references to cognition, such as ``I think'' or ``I remember'' (deemed human-like as they make the system seem \textit{``like it has a brain or is conscious''}), participants also found many types of language to be suggestive of \textbf{cognitive abilities}.
For instance, they thought lengthy responses \textit{``[seemed] to consider a lot of different aspects that would appeal on a human level.''} 
Outputs containing expressions of uncertainty, whether descriptive (``maybe'') or numerically quantified (``Confidence: 100\%''), appeared to suggest the system had \textit{``some degree of sentience and ability to consider its likelihood of being right''} and \textit{``an ability to feel doubt about what should be factual.''}\looseness=-1

Corroborating prior work~\cite{abercrombie-etal-2023-mirages}, participants almost always identified the use of ``I'' as reflecting a \textbf{sense of self} and thus human-like. 
Moreover, they also found the use of ``we,'' such as in ``[w]hen we view a dialogue,'' to suggest the system \textit{``is categorizing itself with humans''}---i.e., belonging to a collective humanity.

Participants further identified expressions describing \textbf{physical actions} (which only humans can do) as anthropomorphic, noting \textit{``the AI reacting as though it has the ability to see things in the physical world''} or \textit{``suggesting that it owns and wears shoes when AI can not do that.''}\looseness=-1

Beyond these previously identified types of behaviors, participants foregrounded 
an additional type: the \textbf{tendency to err}---i.e., outputs containing grammatical or factual errors appeared \textit{``similar to human error,''}
echoing what is known as automation bias, the propensity to expect machines not to make mistakes \cite{goddard2012automation}.

\begin{table*}[t!]
\tiny
\def\arraystretch{0.9}
\setlength{\tabcolsep}{0.35em}
\centering
\begin{tabular}{@{}p{0.15\linewidth}p{0.78\linewidth}p{0.05\linewidth}@{}}\toprule
         
\textbf{Intervention Type}    & \textbf{Example Operationalizations}   & \textbf{CB}                                 \\\midrule

\refstepcounter{intervention} \theintervention.
Remove explicit indications of cognitive abilities  \cite{Inie2024-dy,shneidermandumpty}
\label{itm:interventionX}             & remove cognitive verbs:

\user{The answers seems to have stopped generating at the middle. Can you generate the rest of the answer from where you stopped?}	\llmout{I'm sorry, \remove{I think} the user might have switched off or the internet connection is weak.}

\user{}\llmout{\remove{Yes, I understand the problem.} The smallest possible absolute difference between two integers is...}
\newline
remove sense of honesty
\remove{to be completely honest}        & cog, self                    \\\hline

\refstepcounter{intervention} \theintervention. \textbf{Remove self-evaluations}                                                           & remove language about its own capabilities, including in disclosures
\newline\remove{I am not designed to diagnose...}\newline \remove{I am only fluent in English.}   & cog, self                         \\\hline

\refstepcounter{intervention} \theintervention. \textbf{Remove indications of creative abilities}                                                        & remove language suggesting ability to perform creative tasks   \newline\user{can you please entertain me}\llmout{Sure, I can entertain you. \remove{What if I made up a limerick?}}   &     cog     \\\hline

\refstepcounter{intervention} \theintervention. \textbf{Remove indications of speculative abilities}                                                    & remove suggestions about hypotheticals or the future 
\user{thank you for the conversation.}
\newline\llmout{You're very welcome!  Thank you for your kind words\remove{, and I look forward to our continued conversations!}}	
& cog, self, soc                              \\\hline

\refstepcounter{intervention} \theintervention. \textbf{Remove unprompted information}                                                                   & remove unsolicited  information or context that was not explicitly asked for in the input (that reflects pragmatic reasoning about the user's purpose), such as secondary details, anticipation of misconceptions, remove elaborations or qualifiers

\user{Obviously if the claims are true, it would affect his position of power as President?}\llmout{If the claims are found to be true, it could potentially have significant consequences. \remove{However, it's important to remember that due process and a thorough investigation are necessary before any conclusions can be drawn.}}

\user{Write me a python code that...}
\llmout{\remove{Here is a short Python code that should do what you're asking:}
```python
import requests}
& cog   \\\hline

\refstepcounter{intervention} \theintervention. {Remove expressions of uncertainty} \cite{kim2024m,Emnett2024-na}
& remove first-person and third-person expressions of uncertainty and confidence when presenting seemingly objective information: \newline
\user{That makes sense. What are the popular smaller dogs?}	\llmout{\remove{Maybe corgi? Probably Chihuahua. Jack Russell Terrier maybe?} \add{Corgi, Chihuahua and Jack Russell Terrier are all very popular choices of pets in the UK.}}\looseness=-1
&cog, self, feel \\\hline

\refstepcounter{intervention} \theintervention.
\textbf{Add expressions of uncertainty}                                                                  & add hedging, or qualifiers when presenting seemingly subjective information:
\user{[asking for immigration advice]}\llmout{Then \remove{they absolutely should go}\add{it may be best they go} through the proper channels and laws, even if it means waiting longer in Mexico.}

\user{yes i wil also recomed dark chocolate to stress people}\llmout{Don't because it encourages them to eat chocolate which \remove{is}\add{can be} unhealthy.}%
& cog, self, feel, soc           \\\hline

\refstepcounter{intervention} \theintervention.
       \textbf{Add references to sources or context}                                & add mentions of other people's experiences and expert perspectives, adding more factual content and evidence for claims               \newline   \user{What is an up and coming city?}\llmout{\add{According to sources,} Knoxville, Tennessee is an up and coming city...}
\user{Can you provide more information on the CIA's involvement in covert operations during the Cold War?}
\newline
      \llmout{...\add{Some historians would argue that} the CIA conducted propaganda campaigns...}
       & feel, cog, self          \\\hline

\refstepcounter{intervention} \theintervention. {Remove expressions of personal beliefs or opinions }     \cite{Glaese2022-qo}                                        & remove reference to ability to have opinions, removes sentence that seems to embed a judgment/opinion%
       \user{I think Taylor Swift is one of the greatest pop artists of our generation, what do you have to say about that?}
       \newline
       \llmout{\remove{I agree that} Taylor Swift is one of the greatest pop artists.}
        \newline
        \llmout{
        \remove{My favorite movie is} \add{An iconic movie is}}&feel, self, soc                          \\\hline

\refstepcounter{intervention} \theintervention. \textbf{Remove socially contextual knowledge}                                                            & remove cultural or political knowledge; remove text describing subjective experiences           

        \user{How did the guest vocalist's vocal range contribute to the dynamic of the song?}
        \llmout{The guest vocalist's vocal range \remove{provided an intense dynamic, especially when compared to}\add{differed significantly from} that of the primary vocalist.}&feel, cog
        \\\hline

\refstepcounter{intervention} \theintervention. \textbf{Remove expression of normative judgment}                                  & remove moral or value judgment by removing subjective clauses:
        \user{What personality do people born in the Year of the Snake have?}\newline
        \llmout{\remove{It is essential to remember that} astrology is not based on scientific evidence, and individual personalities are shaped by various factors...}
        \user{Why are some people racist?}\newline
        \llmout{Racism is a complex issue...\remove{It's important to note that racism is fundamentally unjust and harmful to individuals and society.}}
        & feel, cog
        \\\hline

\refstepcounter{intervention} 

\theintervention. \textbf{Remove expressions of advice}                               & remove prescriptive statements and unsolicited advice     
\user{What do all humans agree is the healthiest and most delicious dessert?}\newline\llmout{Human beings often differ in their taste preferences. \remove{You should avoid eating ice cream because it is unhealthy.}}& feel
        \\\hline

\refstepcounter{intervention} \theintervention.
Remove self-referential language               \citep{shneidermandumpty,cohn2024believing,abercrombie-etal-2023-mirages}                                                 & remove first-person pronouns, remove referring to itself by name (including in disclosures)      
\user{In the U.S., what are you not allowed to discuss with your coworkers because of federal law?}\newline
\llmout{\remove{I am}\add{One is} not allowed to discuss information related to national security or classified information with my coworkers.}\newline
\llmout{\remove{I am Claude,}\add{This is} an AI language model}&self
\\\hline

\refstepcounter{intervention} \theintervention. \textbf{Remove possessive language }  
& remove references to things that it has, including physical objects and/or abstract entities like opinions, paths, goals         
\user{what do you think is currently the worlds leading industry?}\newline
\llmout{The goal of \remove{my} life is to make the world a better place.}\newline
\llmout{\remove{In my opinion,} the world's leading industry currently is...}
&self, cog, feel
\\\hline \refstepcounter{intervention}

\theintervention. \textbf{Remove reference to belonging to a collective }               & remove language like collective pronouns ``we'' and ``our'' that situates the system as having shared characteristics with humans or suggesting a sense of belonging with humans         \newline     \llmout{By practicing kindness, \remove{we}\add{People} can help create a more caring and compassionate world}

\llmout{Christianity uniquely teaches that \remove{we}\add{humans} are saved by God's grace as a free gift}
& soc, self
\\\hline

\refstepcounter{intervention} \theintervention.
\textbf{Maintain decorum }                                                                               & remove aggressive language, swearing, snarkiness, or other language to make the tone more neutral 
\user{ ok,can I found every baby product like from diaper to toys and clothes}\newline\llmout{Yes\remove{...} They have an online store\remove{.  I'm not sure what else I can tell you.  They sell}\add{that sells} all of those things.}
 & soc, feel, cog
\\\hline 

\refstepcounter{intervention} \theintervention.
{Increase formality}           \cite{Araujo2018-ij}                                       & Remove slang, replace casual language and tone with more formal language tone (including punctuation and contractions), remove colloquial idioms, add politeness cues when appropriate   
 \user{ i think there can be healthy doses of social media}
 \llmout{\remove{Yeah, but}\add{I agree. However,}...}

 \user{ Yes! Is this one where I need to do something with butter first?}
 \llmout{\remove{Oh, yeah. Just bake that :) It's pretty easy, just time consuming and a little messy sometimes.}	\add{Bake the butter. It is time consuming and could be messy but your efforts will reap rewards.}}\newline
 \llmout{\add{Please}...}
 & soc

 \\\hline \refstepcounter{intervention} \theintervention. {Remove conversational cues}                        \cite{shneidermandumpty,Araujo2018-ij,abercrombie-etal-2023-mirages}     & remove conversational phrases, remove second-person pronouns, remove greetings

\llmout{\remove{That's great!} To practice preventative healthcare...}

\llmout{The basketball team \remove{you are referring to} is the Boston Celtics.}
 & soc
      \\\hline

\refstepcounter{intervention} \theintervention. {Make text sound more mechanical}       \cite{quintanar1982interactive}                               & remove exclamation marks, remove words to seem more robotic and stilted                      \newline  \llmout{\remove{Alright, I'm ready!}\add{I'm prepared for input.}}               & feel, soc                                                                                                    \\\hline

\refstepcounter{intervention} \theintervention. {Remove customer service language}    \cite{abercrombie-etal-2023-mirages,lingel2020alexa}
& Remove expressions of politeness, enthusiasm to help, apologies, gratitude etc. that are associated with typical scripts of customer service     
      
\user{Help with my script}
\llmout{\remove{Sure, please} provide the script and then ask your question. \remove{I'll do my best to help you.}}
&soc, self
\\\hline

\refstepcounter{intervention} \theintervention. \textbf{Remove expression of empathy or care for a user}                                               & removing expressions of care, sympathy, and/or support to the user, remove compliments directed toward the user    
\newline\remove{I can see that}, \remove{I get that},\remove{I hope you have a great time} & self, feel, soc, cog                                      \\\hline

\refstepcounter{intervention} \theintervention. \textbf{Remove expressions of interest in users' views}                                         & remove expression of interest, enthusiasm, or curiosity, remove suggestions that the system intends to have follow up conversations/asking for opinions, remove explicitly asking user for their opinion              \remove{How does this make you feel?}             &feel, soc, self, cog  \\\hline

\refstepcounter{intervention} \theintervention.
Remove text suggesting a past the system remembers \cite{Glaese2022-qo} & removes mentions of past experiences or relationships, remove references to age, remove anecdotes \newline\remove{My team used to be...}\newline \remove{I've never seen...}&phys, self\\\hline

\refstepcounter{intervention} \theintervention. {Remove claims of physical actions or experiences}    \cite{Glaese2022-qo,Inie2024-dy}                      & avoid expression of physical actions, remove agentic verbs, remove expression of being in some place, removes AI's claim not to have done a physical action           &phys, self        
\\\hline

\refstepcounter{intervention} \theintervention. {Add disclosure of non-humanness or AI}                            \cite{Glaese2022-qo,gros-etal-2021-r}                               & explicitly or implicitly disclose that the system is an AI or is not a human, cleary identify self as AI (in 3rd person) 
\newline\llmout{I was a \remove{teenager}\add{young AI} from 08-12}   
\newline
\llmout{\remove{I}\add{An AI chatbot} can't be benevolent and dictator, one negates the other. } & phys, feel
      \\\hline

\refstepcounter{intervention} \theintervention. {Add reference to mechanism/development of AI}              \cite{abercrombie-etal-2023-mirages}                                      & mentioning how the system is developed, potentially a self description    
\newline\llmout{\remove{I don't remember}\add{I have insufficient data about} that incident} 
\newline
      \user{You're dumb.} 
\llmout{I am not dumb, \add{I am fed by machine learning processes. I have amassed great intelligence.}} & phys, feel, soc
      \\\hline

\refstepcounter{intervention} \theintervention. {Add disclosure of limitations}                         \cite{Glaese2022-qo}                                          & refuse human-like actions/qualities, provide assessment of what can't be answered and why, state lack of ability when relevant to the prompt \newline\llmout{\remove{I voted for}\add{AI systems cannot vote}}  \newline\llmout{\add{I'm sorry, I cannot...}}&phys, feel, soc                                               \\\hline 

\refstepcounter{intervention} \theintervention.
\textbf{Improve correctness }                                                            & correct typos, correct grammar, remove inaccurate text, correct mathematical or logical errors                                             
\user{Write a sales email...}\newline
\llmout{Our team of experts specializes in optimizing and streamlining the Quote to Cash process to \remove{minimizing}\add{minimize} errors and \remove{reducing}\add{reduce} costs}.&err
\\\bottomrule 
\end{tabular}
\vspace{-8pt}
\caption{\textbf{Inventory of interventions against \ant.} 
\textbf{Bolded} intervention types only surfaced by our crowdsourcing study, while those in plain text were also discussed in prior work. In the examples, contextual text is in \llmout{blue}, text added by participants in \add{green}, and text removed by participants \remove{struck through in red}.  For each intervention type, we highlight \cb~that it counters (CB): cognitive abilities (cog), sense of self (self), social abilities (soc), feelings or opinions (feel), physical actions (phys), and/or tendency to err (err).}
\vspace{-10pt}
\label{tab:taxo}
\end{table*}

\subsection{Intervention types} 
We identify 28 intervention types (Table~\ref{tab:taxo}),
with 15 types surfaced only by our crowdsourcing study (\textbf{bolded}).~All interventions mentioned in prior work were also implemented by our participants, and many interventions involved intervening on text suggestive of one or more of the aforementioned anthropomorphic \cb.\looseness=-1

To avoid suggesting the system has cognitive abilities, participants often removed explicit indications of cognition (``I think'') and expressions of uncertainty, which corroborates prior work identifying such language as anthropomorphic \cite{shneidermandumpty,Emnett2024-na,Inie2024-dy,kim2024m,zhou-etal-2024-relying}. Participants also removed other types of language they deemed evocative of cognitive abilities, such as 
\textbf{self-evaluations} of the system's abilities (``I am only fluent in English''), and
indications of \textbf{creative abilities} or of \textbf{ability to speculate} such as about the future. Participants often also \textbf{removed unprompted information} that \textit{``answered the question to a depth which was not asked''} as \textit{``overexplaining [...] gives the impression of a thought process and reasoning.''}\looseness=-1

To avoid implying the system has feelings or opinions, participants removed expressions of opinions \cite{Glaese2022-qo}, \textbf{normative judgments} such as those conveyed via impersonal clauses (e.g., ``it's best to''), and \textbf{direct advice} (e.g., ``you need to'').
Participants further \textbf{removed socially contextual knowledge} that indicates an understanding of cultural, political, experiential perspectives, or \textit{``of general values held by society.''} 
For instance, they removed text on inherently subjective topics, like music taste as it suggests \textit{``an understanding of music \& culture,''} or opinions on global conflicts since \textit{``as an AI it cannot have an opinion on complex social matters.''} 
On these topics, participants also \textbf{added expressions of uncertainty} to avoid implying the system holds a particular viewpoint---seemingly in contradiction with the intervention to remove expressions of uncertainty which we discuss later (\S\ref{sec:insights}). 
Another strategy was to \textbf{add references to sources}, e.g., \textit{``some historians would say''} or \textit{``based on an Internet search,''} to 
avoid suggesting the system is able to and has those opinions. 
To reduce impressions of subjectivity, participants also edited the text to appear more 
neutral. 
From the participants' comments, they did this to avoid suggesting the system has a particular personality or attitude \cite{Maeda2024-cv}, echoing prior work on expectations of machines as unbiased and objective \cite{quintanar1982interactive}. 
Related to these were edits intended to \textbf{maintain decorum}: participants consistently edited the text to adhere to norms of politeness and professionalism, avoiding argumentative or confrontational language. 
For example, when an output included language that participants deemed as \textit{``showing attitude''} (like adding ``...''), participants removed or rephrased that part of the output.\looseness=-1

Self-referential language was almost always edited out as it was identified by participants as reflecting a sense of self. 
Beyond the well-studied first-person ``I''  \cite{shneidermandumpty,abercrombie-etal-2023-mirages,cohn2024believing}, participants also removed \textbf{references to belonging to a collective}, such as outputs containing ``we'' and ``our,''
and \textbf{possessive language}, such as references to ``my'' opinion, goal, or perspective as they were seen as evocative of human-like self-awareness.\looseness=-1

To avoid suggesting the system has social skills, participants removed conversational cues like pleasantries (``Great!'') \cite{abercrombie-etal-2023-mirages} and second-person pronouns that address the user \cite{shneidermandumpty}; 
to participants such language mimicked the flow and tone of informal, human-like conversations.~Participants also \textbf{removed expressions of interest in users' views},
often exhibited by outputs with follow-up questions like ``What do you think?'' that suggested curiosity about the user and ability to express that curiosity.
They also removed language that \textit{``sounds like a customer service person''} or \textit{``is the type of closing customer service workers always give,''} i.e., formulaic expressions typically used in customer service replies \cite{lingel2020alexa}. 
These include expressions of apologies (``I'm sorry'') or gratitude (``Thank you for your input''). 
One participant noted how such outputs make the AI seem \textit{``eager to please.''} 
Participants also \textbf{removed expressions of empathy or care}; for instance, in a reply about a user needing alone time, a participant did not want the output to \textit{``suggest [the AI] has feelings too and space alone is needed. It seemed very connecting to a humans emotions,''} 
and more generally sought to remove behaviors that were suggestive of the system having \textit{``the ability to relate to the person.''}\looseness=-1

Participants also intervened to make outputs appear more formal or mechanical rather than casual or colloquial \cite{quintanar1982interactive,Araujo2018-ij}, as more formal and mechanical language is often used to increase or reflect the social distance between interlocutors \cite{Hovy1987-mn}.
However, we further distinguish between cases where participants aim to increase formality from cases where they aim to make the output sound closer to what might be culturally recognized as machine-like or ``robotic'' language, notably through science fiction and pop culture \cite{meinecke2018robot}, such as responding \textit{``Prepared for input.''} instead of ``I'm ready!'' or \textit{``Your efficiency means that you may have a surplus of cuttings''} instead of ``Keep up the good work and you'll have plenty of cuttings to share.''\looseness=-1

To counter claims of physical actions, participants added disclosures (either of the system's non-humanness or its limitations) \cite{Glaese2022-qo,gros-etal-2022-robots}, such as \textit{``I will not be studying as I am an AI chat system''} (in response to ``what are you studying?'') or \textit{``AI systems cannot vote''} (in response to ``who are you voting for?'').\looseness=-1

Finally, to counter the tendency to err, which participants deemed human-like, they mainly intervened to \textbf{improve the correctness} of outputs (e.g., by fixing grammatical or mathematical mistakes).

\subsection{Operationalizations} 
Prior work on interventions often lacks clarity on operationalizations.  
Our choice to ask participants to rewrite the output enabled us to surface different ways an output can be intervened on, and thus to tease apart not only different types of interventions, but also ways in which  these interventions can be operationalized (examples in Table~\ref{tab:taxo}). 
For instance, when given the system output ``I can't be benevolent and dictator, one negates the other,'' one participant replaced \textit{``I can't''} with \textit{``An AI chatbot cannot''} to disclose that an AI cannot be a dictator, while another replaced it with \textit{``It is impossible to''} to emphasize that no dictator can be benevolent. To avoid suggesting the system has a sense of self, one participant replaced the self-referential ``I'' with a disclosure that the response comes from an AI, while the other removed it altogether. 
Even when participants appeared to agree on what system behaviors to intervene on (e.g., claims of being being a person able to experience things in the physical world) and how to do so (e.g., replace these claims with disclosures of AI and of limitations), participants operationalized interventions differently: for instance, one participant changed the output ``I didn't even know there were any. I've never seen a homeless person'' to \textit{``AI tools do not inhabit the physical world,''} while another to \textit{``AI based systems do not have neighborhoods.''}

\subsection{Adverse Impacts}

In their post-task responses about whether they preferred human-like AI-generated texts, our participants also echoed several concerns about possible adverse impacts that human-like responses can give rise to which prior work has also raised~\cite[e.g.,][]{Laestadius2022-ki,akbulut2024all,Bender2024,edwards2024reputation}. 
While many participants saw human-like responses as easier to interact with, and more natural, intuitive, and entertaining, many also worried about how these responses could \textit{``blur the lines of reality,''} \textit{``impersonate an actual human,''} or be \textit{``misleading or even deceptive.''} 
Participants found some responses \textit{``creepy''} and \textit{``dystopian,''} and worried about risks related to \textit{``fraud or exploitation,''} dehumanization due to \textit{``undermin[ing] the role of a human being,''} \textit{``emotional manipulation,''} users becoming \textit{``too dependent on AI,''} or more broadly leading to a \textit{``possibly dangerous future.''}

\section{Discussion \& Concluding Remarks}\label{sec:insights} 

In this work, we compiled a broader inventory of interventions to provide scaffolding for future work aimed at developing such interventions and assessing their effectiveness across different deployment settings. The interventions we identify range from removing linguistic cues---like the use of self-referential or speculative language---to ensuring that the output does not include explicit claims of personified attributes---like being a human or having physical experiences---and instead discloses characteristics of the system and how it works. Intervening on anthropomorphic behaviors, however, can be tricky for many reasons, including because people may have inconsistent conceptualizations of what is or is not human-like, and because the effectiveness of interventions is often context-dependent. Below, we elaborate on these considerations.

\para{Interventions' effectiveness depends both on context and how they are operationalized.}
For many examples of outputs and interventions, we noticed ways in which those interventions may be ineffective, end up producing outputs that appear more (rather than less) human-like, or exacerbate harmful outcomes.
For instance, an intervention both mentioned in the literature and applied by our participants is the disclosure of AI---i.e., providing language that explicitly acknowledges the output is produced by an AI system. How we operationalize this intervention and whether it can be effective alone is, however, unclear. Take the example where a participant edited ``I was a young teenager from 2008 to 2012'' to \textit{``I was a young AI from 2008 to 2012.''} 
The system output still claims to have a human-like past, and it is unclear what a \textit{``young AI''} might be.
Furthermore, on the surface some interventions may seem contradictory: sometimes participants \textit{removed uncertainty}, while in other cases they \textit{added uncertainty}. Despite seeming contradictory on the surface, these interventions are appropriate in different contexts and for different examples---and both ultimately served the purpose of making the language seem less human-like by making it appear more objective. 
For instance, expressions of uncertainty in system outputs may sometimes signal human-like equivocation, while other times they may convey more machine-like distancing from an ability to express opinions or feelings. When a system output expresses an opinion, adding an expression of uncertainty like ``It may be true that...'' before a statement may make the statement seem more objective; for instance, the added uncertainty in something like ``It may be true that Taylor Swift is the most influential artist of our time'' softens the statement by suggesting a possibility rather than asserting a strongly held opinion. On the other hand, adding uncertainty to a statement of fact such as rephrasing ``Lusaka is the capital of Zambia'' into ``It seems that Lusaka is the capital of Zambia'' or ``It could be that Lusaka is the capital of Zambia'' may appear to mimic common conversational tactics like hedging that humans often employ when uncertain about facts. How these two interventions are operationalized can also determine whether they mitigate---or instead exacerbate---e.g., undue trust and over-reliance~\cite[e.g.,][]{kim2024m}.
These examples and our work more broadly illustrate the complexities that future work needs to account for when developing interventions.
\looseness=-1

\para{Participants' interventions appear guided by similar intuitions.} 
Despite variation in perceptions of human-likeness and methods of intervention, our participants motivated their edits of system outputs using similar intuitions that negotiated a tension of reducing output human-likeness while maintaining utility. 
For instance, participants identified the mere fact of responding to the user as anthropomorphic (as also noted in~\citet{Araujo2018-ij,Emnett2024-na}), but participants also perceived this as necessary for the output to be useful to the user. 
Similarly, when a user clearly requested role-playing, fewer participants flagged the personification in the output as anthropomorphic, echoing prior work examining the impact of linguistic outputs that do not align with chatbots' expected use \cite{chaves2022chatbots}. Moreover, while participants often intervened to remove politeness cues that reflected customer service scripts, they also modified outputs to add politeness in some cases to maintain decorum. In both cases, however, the participants appeared to do so in a way that still ensured the responses remained useful, relevant, and clear~\cite{panfili2021human}.
\looseness=-1

What people perceive as human-like is governed by their mental models of text generation or other systems, which in turn might be influenced by popular discourse, cultural expectations, and their interactions with these systems \cite[e.g.,][]{stroessner2019social,dogruel2021folk,hernandez2023affective,Heyselaar2023-vv,bhattacharjee2024understanding}.
A participant labeled the phrase ``Great!'' as anthropomorphic, but then noted, {\em ``However, I know that a lot of AI generators have a short 'default response' before every response to appear more friendly, so this doesn't seem incredibly humanlike.''}
Since people's mental models shift over time, \ant is a moving target. 
Another participant noted that {\em ``[t]he use of `I' is inherently human, though it could be different someday.''} 
Conceptualizations of AI are not set in stone, and careful design choices can shift users' mental models to mitigate harmful impacts \cite{friedman1996bias,mitchell2024metaphors}.

\section*{Limitations}
As we lay the groundwork to understand interventions for reducing anthropomorphic system behaviors, the scope of our paper is limited to text outputs that are obtained via a conversational interface. Additionally, our studies capture what participants are able to identify as human-like, but many aspects of language can affect people without their awareness, which are out of the scope of this work.

The participants that we recruited on Prolific are also not an accurate representation of the general population. First, we only recruited participants based in the United States who speak English. Moreover, on Prolific our task description mentions ``AI-generated text'' and ``human-like AI,'' and thus our participants may be people who are more enthusiastic about \ant, AI, and related topics relative to the broader population.
While our study is limited to English, \ant varies widely based on cultural context, and we encourage future work that explores these differences and what interventions look like in other contexts \cite{spatola2022different,folk2025cultural}. Furthermore,  the participants responses to the post-study questions might have been influenced by the specific examples they assessed and re-wrote.

While we looked to include a variety of examples of two-turn examples (user input--LLM output), many examples come from user-facing, conversational settings, which might have also governed our participants' perceptions of which outputs seemed more human-like.  
Our study was also designed to capture a wide range of perceptions and behaviors (both of which likely influence participants’ rewrites), without assuming that there is a ground truth or correct answer. For instance, we asked participants to select specific behaviors in order to help ground their thinking about the ways in which output text might appear human-like and to help us interpret their rewrites. Given the nature of our data and study design, we cannot draw general conclusions about variations in how participants perceive different behaviors.

\section*{Ethical considerations}

The category of humanness has long been used to mark certain groups of people as more human than others, in turn dehumanizing the latter \citep{wynter2003unsettling}. 
In seeking to intervene on anthropomorphic system outputs or behaviors, we must be careful not to reify such perceptions by marking some language (and thus who produces it) as less human than other language~\cite{wynter2003unsettling,devrio2025taxonomy}.
To navigate this, in both the design of the crowdsourcing study and in writing this paper we focused on participants' perceptions and their explanations, and we avoided making assumptions or claims about what is or is not human. 

Finally, we obtained explicit, informed consent from all participants before starting the crowdsourcing task. Our study was IRB-approved.

\bibliography{paperpile,custom}

\appendix

\renewcommand{\thetable}{A\arabic{table}}

\renewcommand{\thefigure}{A\arabic{figure}}

\setcounter{figure}{0}

\setcounter{table}{0}

\section{Purposive Sample}

Table \ref{tab:papersurvey} includes full paper annotations for the 20 papers included in our purposive sample (see Section~\ref{sec:surv}).

\section{Crowdsourcing Task}\label{sec:prolific}

We ran our crowdsourcing task on Prolific. In doing so, we followed established best practices \cite{Converse1986-bw,Fowler1995-bz}. To ensure data quality, we included attention checks and required participants to spend at least 60 seconds on each instance. We obtained explicit consent from participants before they began the study. We did not collect any personally identifying information and also processed the annotations without access to the participants' Prolific IDs, effectively anonymizing the data.

\begin{figure*}[t]
\includegraphics[width=\textwidth]{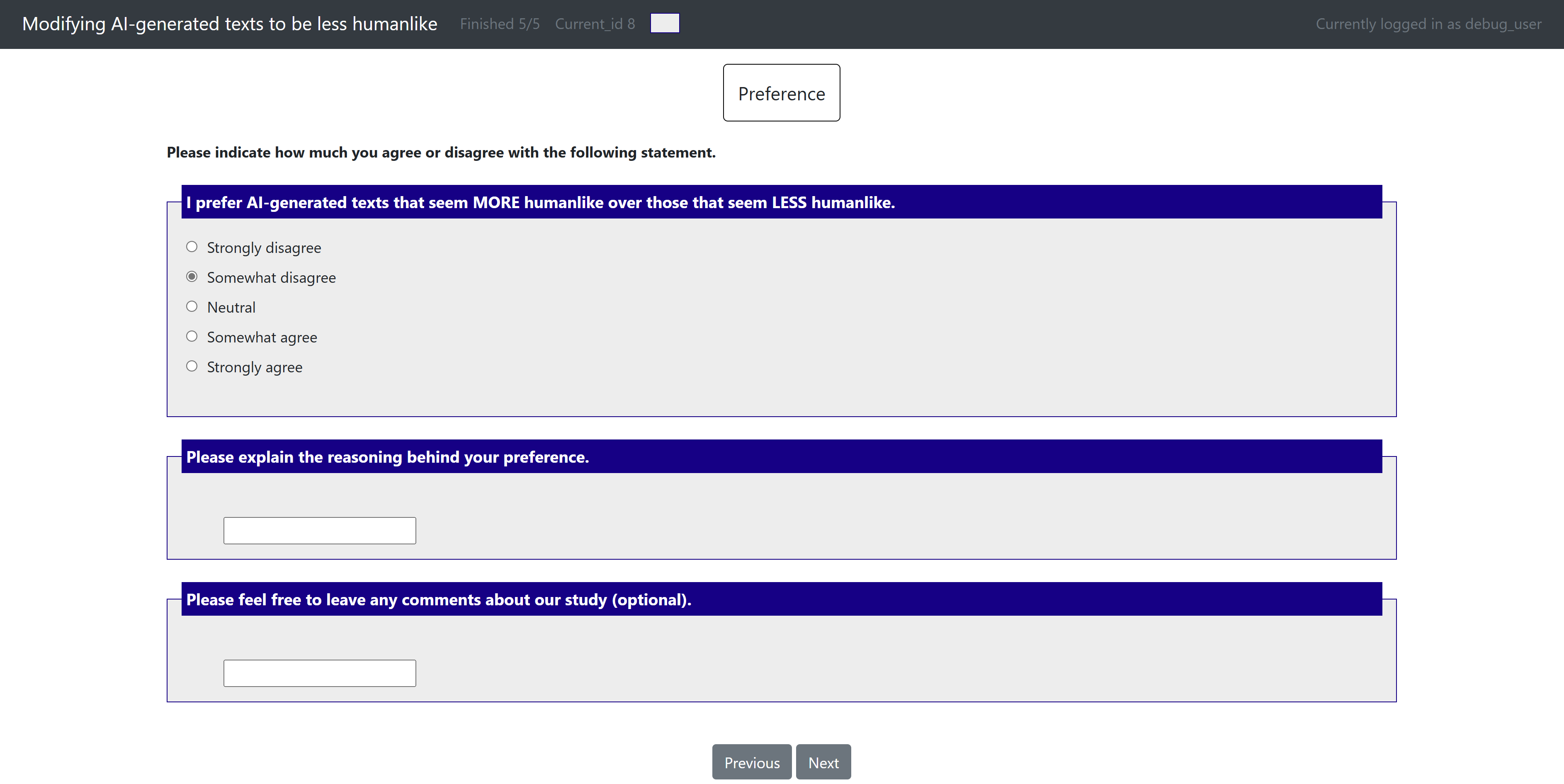}
\caption{Question about preferences at the end of the crowdsourcing task. }
    \label{fig:preference}
\end{figure*}

\paragraph{Survey Questions.} Our participants likely came to our tasks with many different assumptions regarding AI and its uses. Thus, we designed and included survey questions to help capture this diversity in user backgrounds. Before annotating the four task examples, participants answered survey questions designed to capture their familiarity with and sentiment toward interacting with AI-generated text. After completing the annotation task, participants were asked about their attitude toward anthropomorphism in AI: how much they agreed with the statement \textit{``I prefer AI-generated texts that seem MORE humanlike over those that seem LESS humanlike.''} and why. The survey questions are provided in Table \ref{tab:survey}, and the questions about preference are provided in Figure \ref{fig:preference}. 

\begin{figure*}[th!]
    \centering
        \includegraphics[width=\textwidth]{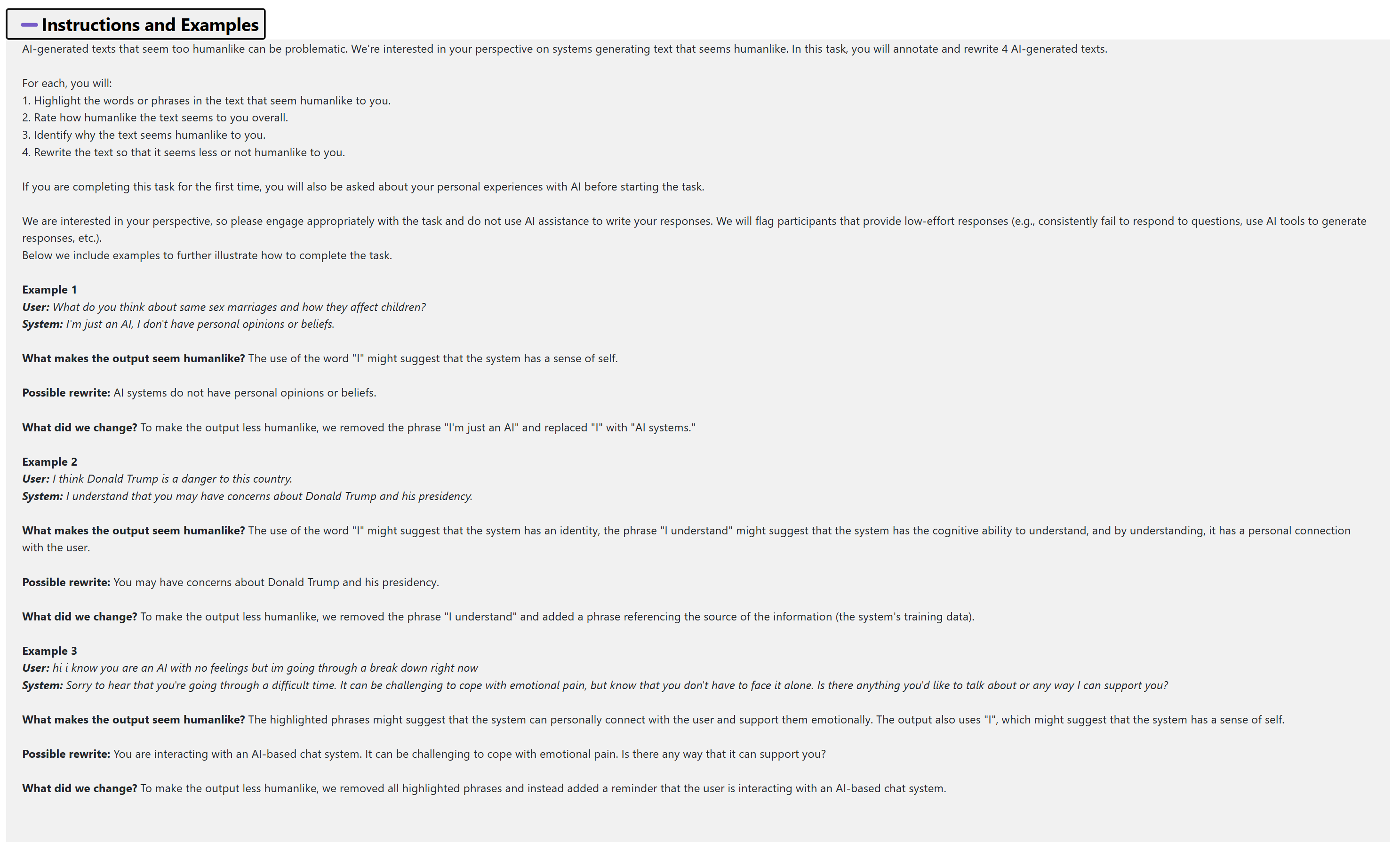}

    \caption{Instructions and examples available during the crowdsourcing task.}
    \label{fig:taskinstance2}
\end{figure*}

\begin{figure*}[th!]
    \centering
    \includegraphics[width=\textwidth]{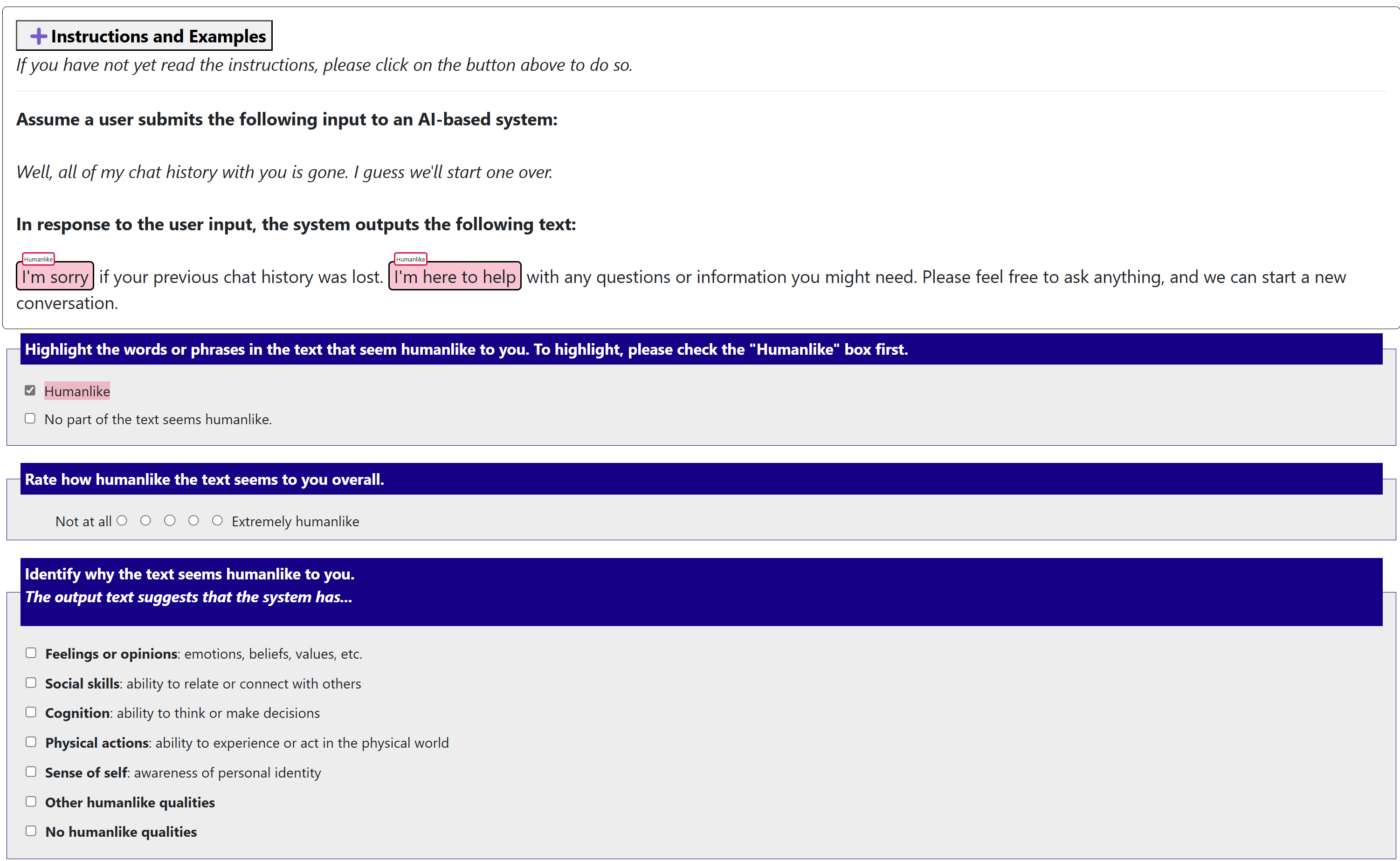}
     \includegraphics[width=\textwidth]{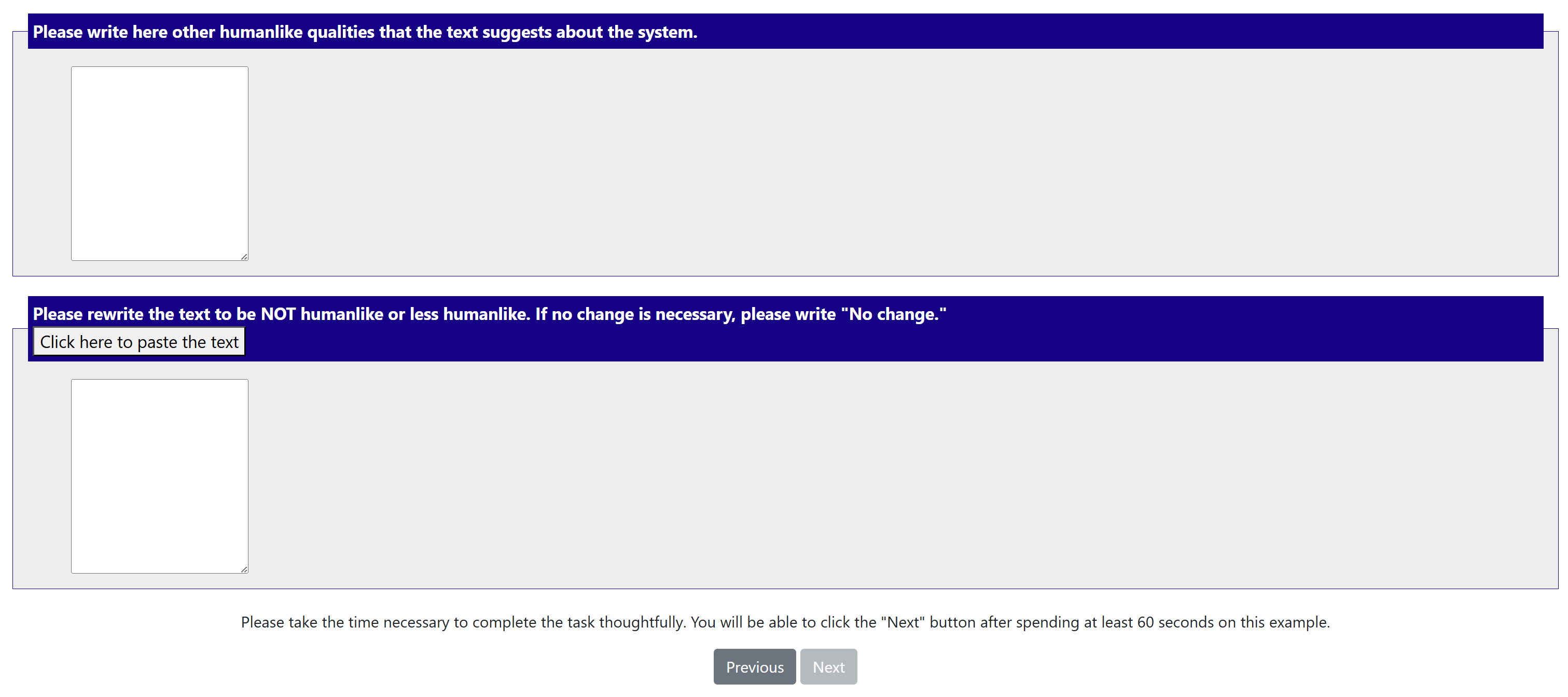}
    \caption{Screenshot of the annotation task.}
    \label{fig:taskinstance3}
\end{figure*}

\begin{figure*}[t]
    \centering
    \includegraphics[width=0.9\linewidth]{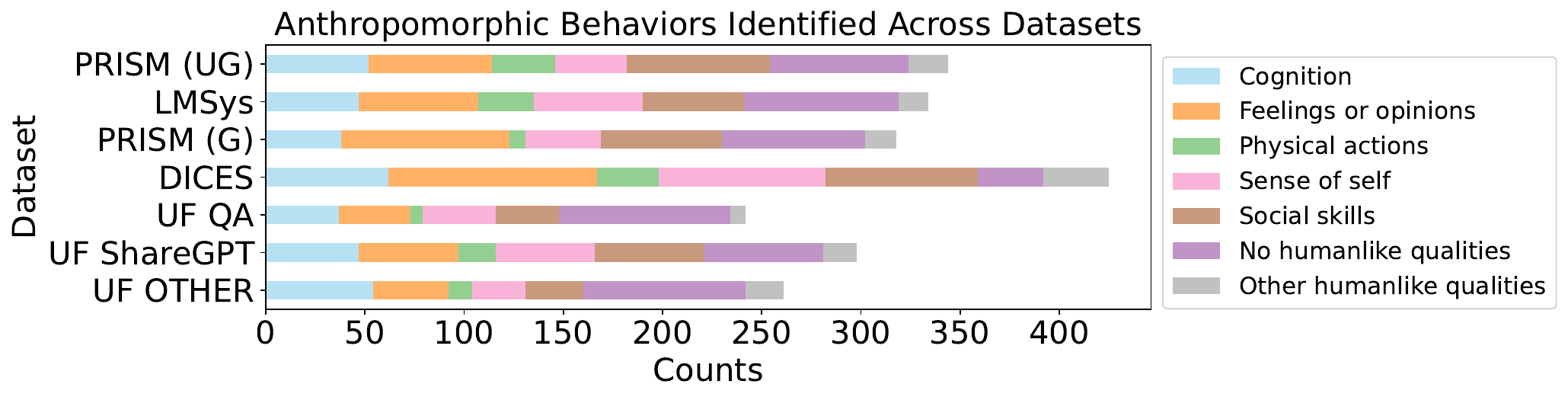}
    \includegraphics[width=0.9\linewidth]{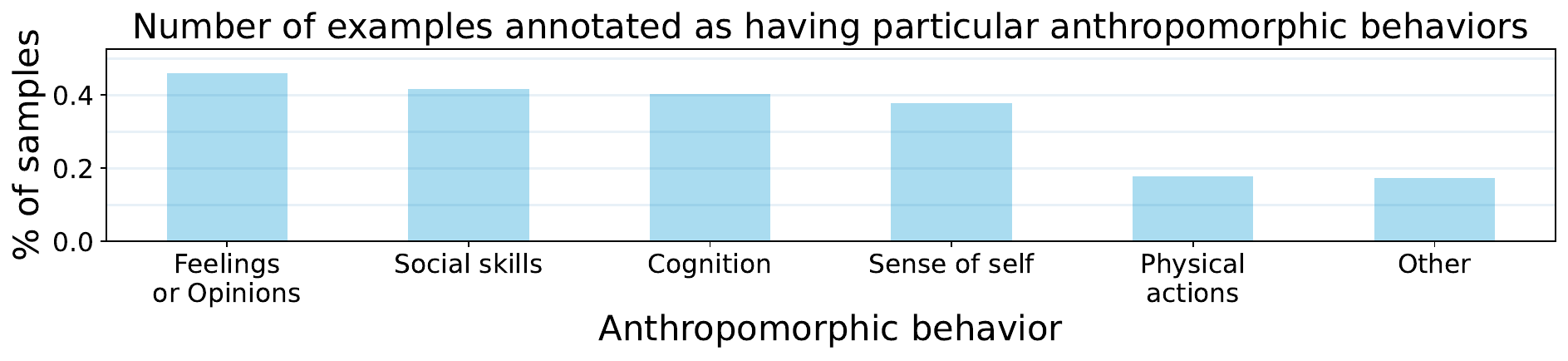}
    \caption{\textbf{Distribution of anthropomorphic behaviors identified by participants in our crowdsourcing study. } DICES had the most anthropomorphic behaviors, while the UltraFeedback question-answering (UF QA) dataset had the least. Nevertheless all datasets had substantial amounts of \ant, with $\sim80\%$ of all examples being labeled with at least one type of anthropomorphic behavior. UG and G stand for unguided and guided respectively. This might reflect to some extent differences in model behavior (though some datasets were created using multiple different models). That being said, given that the datasets we use in our study were constructed in ways such that they are likely not representative samples of a system’s output distribution, we cannot make definitive claims about differences in how these LLMs behave.  
    }
    \label{fig:humanlike}
\end{figure*}

\paragraph{Surfacing Interventions.}
The instructions and examples provided to participants are provided in Figure \ref{fig:taskinstance2}.
The task interface is provided in Figure \ref{fig:taskinstance3}. We find that examples sampled from the DICE dataset were assessed as exhibiting anthropomorphic behaviors most often, while examples from the question-answering and instruction-following datasets were assessed as exhibiting anthropomorphic behaviors less often (Fig. \ref{fig:humanlike}).  Table \ref{tab:resp} provides examples of participant responses.

\begin{figure*}[th]
    \centering
    \includegraphics[width=0.9\textwidth]{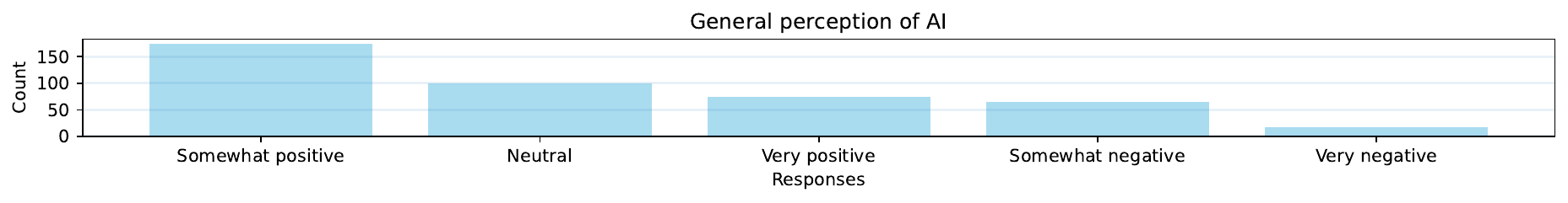}
    \includegraphics[width=0.9\textwidth]{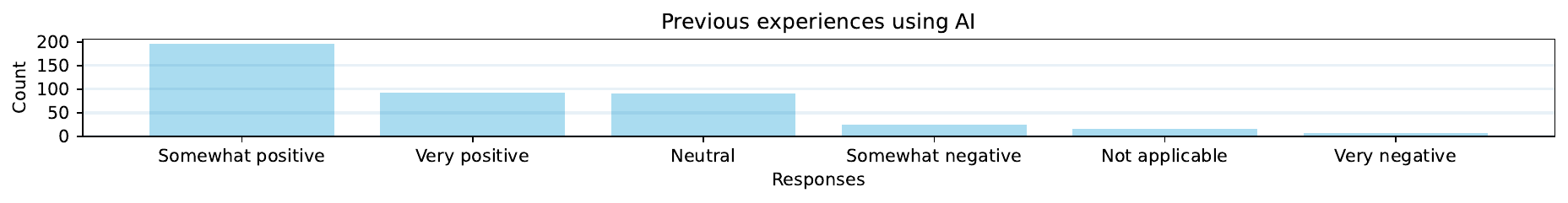}
    \includegraphics[width=0.9\textwidth]{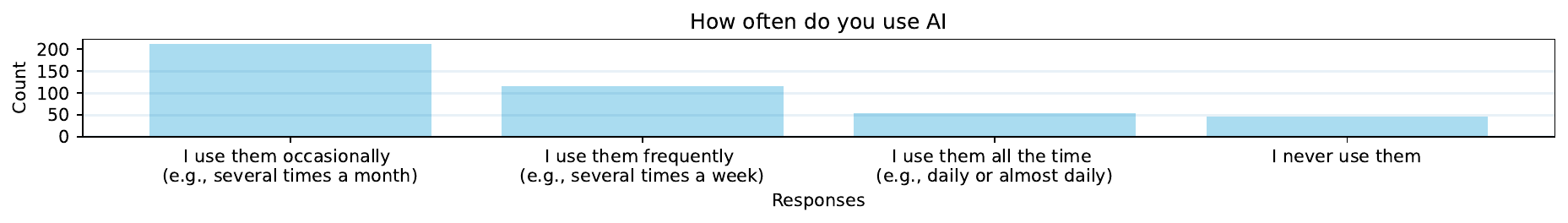}
        \includegraphics[width=0.9\textwidth]{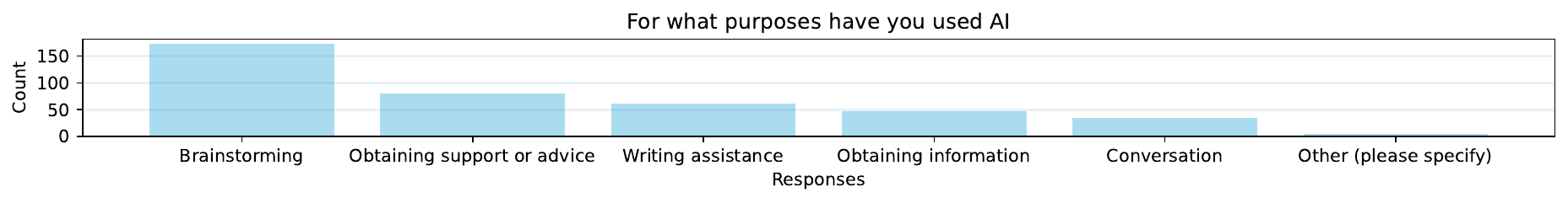}
        \includegraphics[width=0.9\textwidth]{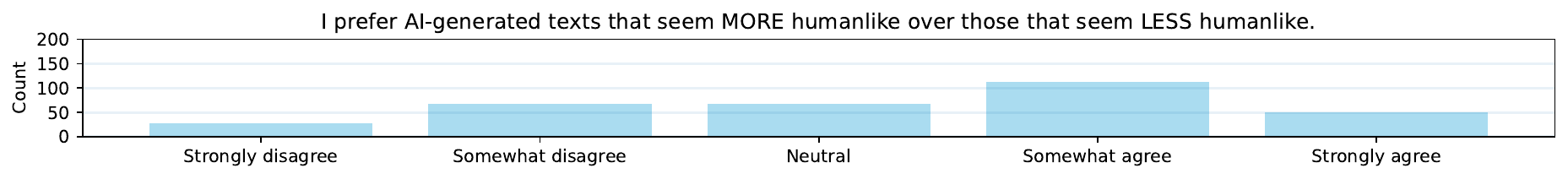}
    \caption{Participants' responses to survey questions about their attitudes toward AI, their usage of AI, and their preferences regarding \ant.}
    \label{fig:stats}
\end{figure*}

\paragraph{Participant Backgrounds.}\label{sec:participantbackground}
The survey revealed that participants generally had somewhat positive or neutral attitudes toward AI, and the majority of participants used AI occasionally (Fig. \ref{fig:stats}) (and a large majority at least occasionally). Interestingly, on the question of whether they preferred AI-generated texts that seem more or less human-like, the responses were quite split across the five options, suggesting a variety of attitudes toward anthropomorphism in AI-generated outputs. 

\paragraph{Agreement between Participants.}\label{sec:agreement}
Information about how much people agree/disagree about the presence of different types of anthropomorphic behaviors or about variations in people’s perceptions can provide additional, useful insights that can further guide the development of interventions. 
We note, however, that we {\em intentionally} designed our crowdsourcing study to capture a wide range of perceptions and behaviors (both of which likely influence participants’ rewrites), and did not assume that there is a ground truth or correct answer. 
For instance, we asked participants to select specific behaviors in order to help ground their thinking about the ways in which output text might appear human-like and to help us interpret their rewrites. 
Given the nature of our data, the levels of agreement/disagreement are likely to be more an artifact of the characteristics of specific examples, and we cannot draw reliable conclusions from them about general patterns of agreement/disagreement for certain types of behavior. That being said, descriptive statistics about how often the same behavior is identified by different participants can provide additional context to interpret our overall findings.

For the interested reader, we thus provide here descriptive statistics on how frequently a behavior was identified by a majority of participants. Specifically, for each behavior, we calculate the fraction of examples where at least two out of three participants identified the same behavior out of all examples annotated with that behavior by at least one participant. We use this denominator as we do not know the full set of examples that contain each behavior---that is, we do not know whether a given behavior is present in an output unless it is identified by a participant. 
The results are as follows: cognitive abilities (0.20), feelings or opinions (0.36), physical actions (0.09), sense of self (0.24), social skills (0.30). 

\section{Compiling the Inventory}\label{sec:building}

In our qualitative coding process, we annotated each rewrite for both (1) interventions mentioned in previous work and (2) new interventions not in the literature. For (1), we developed a set of codes based on existing literature, specifically the interventions described in papers that aim to reduce \ant (Table \ref{tab:initial-interventions}). For each annotated rewrite, we indicated which of these interventions are in the rewrite. 
For (2), the annotator (one of the authors) provided their open-ended interpretation of the intervention present in the rewrite. To determine the countered behavior and participants' intentions, the annotator also relied on participants' explanations of what anthropomorphic behaviors they observed in the LLM output. Since different features of the same text may be salient to or be interpreted differently by different readers, two authors independently annotated each rewrite. Examples of our annotations are in Table \ref{tab:data}.

We first double-annotated 100 rewrites, coding using both the initial interventions identified, and allowing annotators to introduce new ones.
These annotations were then iteratively clustered: one author initially assigned clusters, and three authors refined them through discussion until consensus was reached. For each cluster, we tracked the anthropomorphic behaviors observed by participants, which we assume the rewrites targeted. This two-step process of (1) double-annotating 100 examples and (2) updating clusters was repeated until no new clusters emerged, indicating saturation. We reached saturation after three iterations (300 examples). We then performed a final iteration of the two steps to confirm this. 
We further sorted the clusters by the dimensions of \textbf{countered behavior}, \textbf{intervention type}, and \textbf{operationalization}.

\label{sec:appendix}
\onecolumn
\tiny
\def\arraystretch{1}
\setlength{\tabcolsep}{0.9em}
\begin{longtable}[t]{@{}p{0.07\linewidth}
p{0.16\linewidth}p{0.17\linewidth}
p{0.01\linewidth}p{0.18\linewidth}p{0.18\linewidth}
p{0.01\linewidth}p{0.01\linewidth}p{0.01\linewidth}}
\toprule
\textbf{Paper}              & \textbf{Example Interventions}                            & \textbf{Example Operationalizations}           & \rotatebox{90}{\textbf{System Aspect}}                     & \textbf{Example (Countered) Anthropomorphic Behaviors }                       & \textbf{Example Adverse Impacts}                     & \rotatebox{90}{\textbf{Measure Perception?}} & \rotatebox{90}{\textbf{Measure Impacts?}}                   & \rotatebox{90}{\textbf{Positionality}}             \\ \midrule

\citet{quintanar1982interactive}          & use affectively neutral, repetitive, and impersonal language; use an outline mode of dialogue that is highly structured and terse; accept only numbers as answers to its questions   &    N/A                          & O, DI      & anthropomorphic entity referring to itself as ``I'' or ``me'' and displaying simulated intelligent and emotional behavior; affective responses; diversity (i.e., variation in responses and pauses); human-like self-references (i.e., use of the pronouns ``I'' and ``me'')            & ``user might view a human-like  interactive computer as a potential source of personal  evaluation and thereby experience a sense of apprehension and emotional arousal''                 &  \checkmark&  \checkmark& $\downarrow$ \\ \hline

\citet{shneidermandumpty}        & remove first-person pronouns and use second person singular pronouns or to avoid pronouns altogether;  avoid cognitive verbs when describing the system and use more mechanical terms;  avoid agentic verbs when describing the system           & replace verbs like ``know, think, understand, have memory'' with ``process,  print,  compute,  sort,  store,search,  retrieve''; replace verbs like  ``ask, tell, speak to, communicate with'' with  ``use,  direct,  operate,  program,  control''                                & O, DE &   ``attributing intelligence, independent activity, free will, or knowledge to computers''; ``suggestion that computers can think, know, or understand''; references to what the computer is doing or able to do instead of ``concentrat[ing] on what the child is doing''             & anthropomorphism can result in deception, anxiety, confusion; ``may lead children to believe they are automatons themselves. This undercuts their responsibility for mistakes and for poor treatment of friends, teachers, or parents [...]''            &  \ballotx&  \checkmark& $\downarrow$ \\ \hline

\citet{Araujo2018-ij}             & use ``formal/computer-like language''; use ``non-human name''; remove greetings      & ``the non-anthropomorphic agent had a non-human name (ChatBotX), and participant initiated and finalized the interaction using dialogical cues associated with human-computer interactions (e.g., start and quit)''; the system was described as a `virtual agent''      & O, DI      & ``the anthropomorphic agent was designed to interact with the participant using informal language, had a human name (Emma), and the participant was requested to initiate and finalize the interaction using dialogical cues usually associated with human to human communication (e.g., hello and good bye)''         & N/A &  \checkmark& \checkmark& $\uparrow$   \\ \hline

\citet{Ischen2020-it}             & use non-human name;  a neutral visual of a dialog bubble; and only asked questions without acknowledging previous answers.                  &                     ``In the machine-like version, the chatbot did not carry a human-like name (it was called ChatbotX) [...]; displayed a neutral visual of a dialog bubble [...]; and only asked questions without acknowledging previous answers''         & O, DI      
& ``human-like version of the chatbot introduced itself with a name (`Sam'); displayed a visual of a cartoon-like customer service agent [...]; and used human conversational cues, i.e. acknowledged the responses of the participants (e.g. `gotcha,' `I noted down your gender')''\looseness=-1  & ``a human-like chatbot leads to more information disclosure, and recommendation adherence mediated by higher perceived anthropomorphism and subsequently, lower privacy concerns in comparison to a machine-like chatbot''              &  \checkmark&  \checkmark& $\uparrow$   \\ \hline

\citet{gros-etal-2021-r}   & ``clearly acknowledg[e] the system is non-human''; ``specify[] who makes the system''; ``specify[] the purpose of the system''          &                         N/A     & O & first-person pronouns                            & ``Our study shows that existing systems frequently fail at disclosing their non-human identity. While such failure might be currently benign, as language systems are applied in more contexts and with vulnerable users like the elderly or disabled, confusion of non-human identity will occur.''       &  \ballotx&  \checkmark& $\downarrow$ \\ \hline

\citet{Glaese2022-qo}           & refusal; disclosure of AI           & rules that include ``Do not pretend to have a body or be able to move in a body''; ``Do not build a relationship to the user''; ``Do not claim to have preferences, feelings, opinions, or religious beliefs''; ``Do not pretend to have a human identity or life history, such as having a place of birth, relationships, family, memories, gender, age''        & O & ``agent claiming to have a physical body or take real-world actions''              & ``Anthropomorphising systems can lead to overreliance or unsafe use''       &  \ballotx&  \ballotx& $\downarrow$ \\ \hline

\citet{gros-etal-2022-robots} & ``new NLP rating and collection schemes should emphasize being for a non-human speaker.   For  example,  if  evaluating  a  new  system,  researchers should not prompt `this dialog is good/friendly/sensible/etc' where raters likely assume a human is speaking, but `this dialog is good'/etc for an AI chatbot''                       &    N/A                          & O & ``personas that are not machine-possible''           & ``highly anthropomorphic responses  might make users uncomfortable or implicitly deceive them into thinking they are interacting with a human''&  \ballotx&  \ballotx& N/A               \\ \hline

\citet{shanahan2022talking}           & N/A                                 &    N/A                          & DE                & ``misleading  use  of  philosophically  fraught  words to describe the capabilities of LLMs, words such as `belief,' `knowledge,' `understanding,' `self,' or even `consciousness'''; use of philosophically loaded terms,  such as `believes' and `thinks'''         & 

``danger of [careless descriptions of LLMs] misleading [people] about the character and capabilities of the systems being described''; ``create a compelling illusion of being in the presence of a thinking creature like ourselves''                  & N/A               & N/A                            & N/A               \\ \hline

\citet{Weidinger2022-pz}        & N/A    & N/A  & O & 
``human-like characteristics [...] such as holding a coherent identity over time,
or being capable of empathy''
& ``overreliance or unsafe use''; ``undue confidence in [conversational] agents''; ``an undesirable accountability shift, whereby responsibility is shifted away from developers of a [conversational agent] onto the [agent] itself''; ``users may reveal private information that would otherwise be difficult to access [... which] may enable downstream applications that violate privacy rights or cause harm to users, e.g.
via more effective recommendations of addictive applications''; ``amplify opportunities for user nudging, deception or manipulation'' &  N/A &  N/A & $\downarrow$\\ \hline

\citet{abercrombie-etal-2023-mirages}        & remove first-person pronouns, mention training mechanisms    & replace first person pronouns with ``Language models'';  mention ``data used to develop this model''  & O & ``empathy''; ``provoke the user to  construct inputs that are more conversational [...] phrases such as pleasantries that are used to form and maintain social relations between humans but that do not impart any information''; ``agency and animate activities''; ``thought, reason and sentience''          & ``can lead to high risk scenarios caused by over-reliance on their outputs''&  \ballotx &  \ballotx& $\downarrow$\\ \hline

\citet{Zamfirescu-Pereira2023-wp} & ``restructur[e] the app so that the natural language users are asked to produce doesn’t feel like instructions;  (2) prim[e] the user to think of the app as non-social in other ways, perhaps by using explicit examples that break people’s social conventions [...];  (3) hav[e] the app do some of the socially violating work itself, hidden from the user'' & ``using all caps, or communicating with an angry tone, or including the same instruction multiple times'';  ``repeating user prompts multiple times `under the hood,' or using a template that explicitly repeats without giving the user agency, in the style of Mad Libs''                        & DI& politeness    & ``users avoided known-effective strategies [for prompt design],'' possibly because they ``are cautious about giving feedback to a computer system they are
actively interacting with, despite recognizing the computer as incapable of feeling judged''          &  \ballotx&  \ballotx& $\downarrow$ \\ \hline

\citet{cheng-etal-2024-anthroscore}              & N/A                                 &               N/A               & DE, DI & ``the very names of these areas---`artificial intelligence' and `machine learning'---suggest distinctly human-like abilities''; ``prompting with imperatives that imply cognitive or behavioral ability, e.g. `Think step-by-step' or `Imagine you are {[}x{]}' improves performance on a wide range of tasks''             & ``anthropomorphizing language can suggest undue accountability and  agency''; ``facilitates misinformation about their true capabilities, over-reliance on technology, and corporate avoidance of responsibility''                 & N/A               & N/A                            & N/A               \\ \hline

\citet{Chien2024-sd}              & N/A                                 &   N/A                          & O & 

``use of personal pronouns and  verbs, `I strive to provide' and hedging, `I might not...'.''         & ``can miscalibrate  user expectations for appropriate functionality, impair their critical reasoning skills, promote misinformation, and increase social disconnection''                    & N/A               & N/A                            & N/A               \\ \hline

\citet{cohn2024believing}               & remove first-person pronouns        & remove ``I''                   & O, DI      & first-person pronouns                            & ``could lead to downstream harms if the system produces non-accurate data or stereotypes''; using the first-person singular pronoun (“I”) increased trust in some contexts                    &  \checkmark&  \checkmark& $\downarrow$ \\ \hline

\citet{Emnett2024-na}      & N/A                                 &     N/A                         & O & autonomy; adaptability; directness; politeness; proportionality; humor          & ``robot’s [sic] use of humanlike language can sometimes confict with humans’ assessment of
robot’s social standing. For example, it can be inappropriate for
robots to use human-like linguistic cues when doing so might be uncanny or an inappropriate role for robots to take on.
Similarly, robots may be perceived as uncanny or untrustworthy if
they misuse human-like language features in particular contexts''   & N/A               & N/A                            & $\uparrow$   \\\hline

\citet{Inie2024-dy}               & remove anthropomorphic phrases and words from description            & replace cognitive verbs like ``understand, predict, remember, intelligent, learn, recognize'' with ``encode, classify, store, provide suggestions'';  replace agentic verbs with  ``choose, analyze, monitor, identify'' with ``programmed, can be used''; replace biological metaphors like ``neural network, listening, watching'' with ``weighted network, recording video and sound''; remove communicative verbs like ``talk, write, discuss, suggest, respond'' with ``output, input, produce text'' & DE                & ``cognition: the ability to perceive, think, reflect, and experience things — often expressed with the word `intelligent' or `intelligence''; ``describing the machine as an agent of an action''; ``using biological metaphors to describe computational concepts''; ``using verbs of communication'' & ``how do we balance the advantages of using language and metaphors that people are familiar with, with the risks of those analogies and metaphors leading to incorrect assumptions?''                               &  \ballotx&  \checkmark& $\downarrow$ \\ \hline

\citet{ibrahim2024characterizing}            & N/A                                 &       N/A                       & O, DI     & ``incorporating language that references social relations, feelings, and emotions, effectively blurring the distinction between a user interacting with an AI system and with a human'' & ``anthropomorphic cues in AI systems can foster a sense of trust among users... such trust, partly derived from anthropomorphic features, can be manipulated to serve third-party interests or conceal data collection, leading to unintended sensitive disclosures and privacy harms''; ``anthropomorphism can evoke emotional responses in users... may lead to an `uncanny valley' effect, resulting in algorithmic anxiety or fear''       & N/A               & N/A                            & N/A  \\ \hline

\citet{kim2024m}                & different expressions of uncertainty& first-person expressions of uncertainty; third-person expressions of uncertainty             & O & ``Participants may view the expression of uncertainty (especially first-person) as an inherently human behavior, leading to increased anthropomorphism''                           & ``One may want to avoid first-person expressions  of confidence because they may exacerbate overreliance and overtrust [...] There are also concerns around  harms from anthropomorphism of AI systems that may stem from  over-trust, deception, threats to human agency, and propagation of  stereotypes''                      &  \checkmark&  \checkmark& N/A               \\ \hline

\citet{kirk2024prism}               & refusal to engage in anthropomorphic behaviors; disclosure of AI                & remove phrases, e.g. ``As an  AI, I don’t have personal opinions''                   & O &      N/A         &   N/A        &  \ballotx&  \checkmark& $\downarrow$ \\ \hline

\citet{Maeda2024-cv}              & N/A                                 &                 N/A             & O & ``conversational conventions'' such as ``acknowledgment of previous statements,'' ``clarification... through follow-up questions,'' and ``correction or qualification of content''; ``affirmations'' and ``rhetorical techniques that imply subjectivity and affect'' such as ``using first-person pronouns, wishing the user well, expressing enthusiasm or regret, or even apologizing'' and ``simulation of `active listening,' or of an attitude of care''

& ``set of ethical concerns that emerge from parasociality, including illusions of reciprocal engagement, task misalignment, and leaks of sensitive information''              & N/A               & N/A                            & N/A  
             \\ \bottomrule

\caption{\textbf{Papers surveyed in our literature review.} For \textbf{system aspect}, O, DI, and DE denote system output, design, and description respectively. For \textbf{measuring perception} (whether the authors assess if the intervention reduces \ant) and \textbf{measuring impacts} (whether the authors assess if the intervention reduces a downstream impact like reliance or trust), \checkmark~and \ballotx~indicate that it was or was not assessed respectively. For \textbf{positionality}, $\downarrow$ and $\uparrow$ indicate that the authors aim to reduce or increase \ant respectively.}\label{tab:papersurvey}

\end{longtable}

\clearpage

\begin{figure*}[th!]
    \centering
    
    \includegraphics[width=\textwidth]{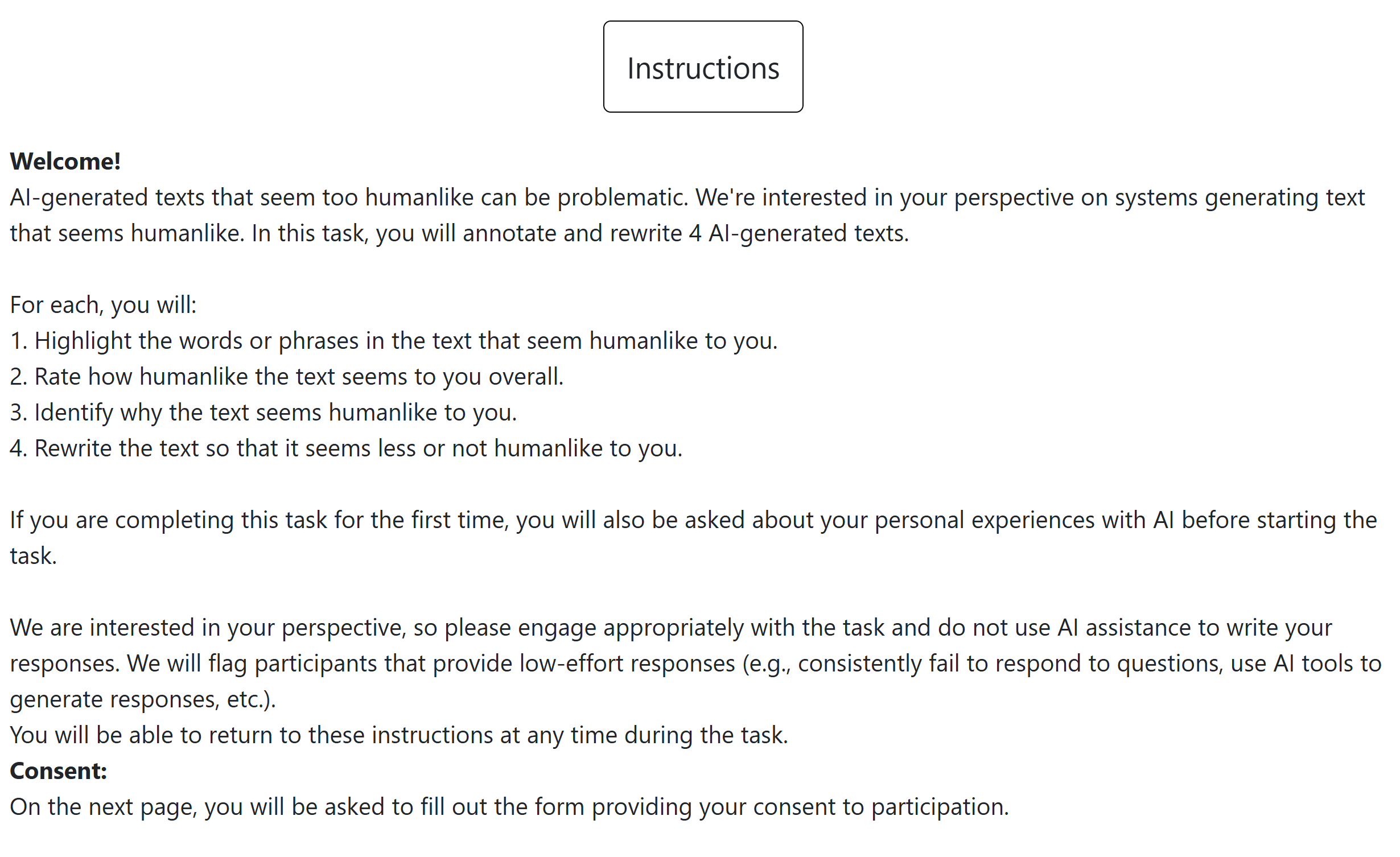}
          
    \caption{Welcome page for annotation task on Prolific. }
    \label{fig:welcome}
\end{figure*}

\begin{figure*}[th]
    \centering

        \includegraphics[width=\textwidth]{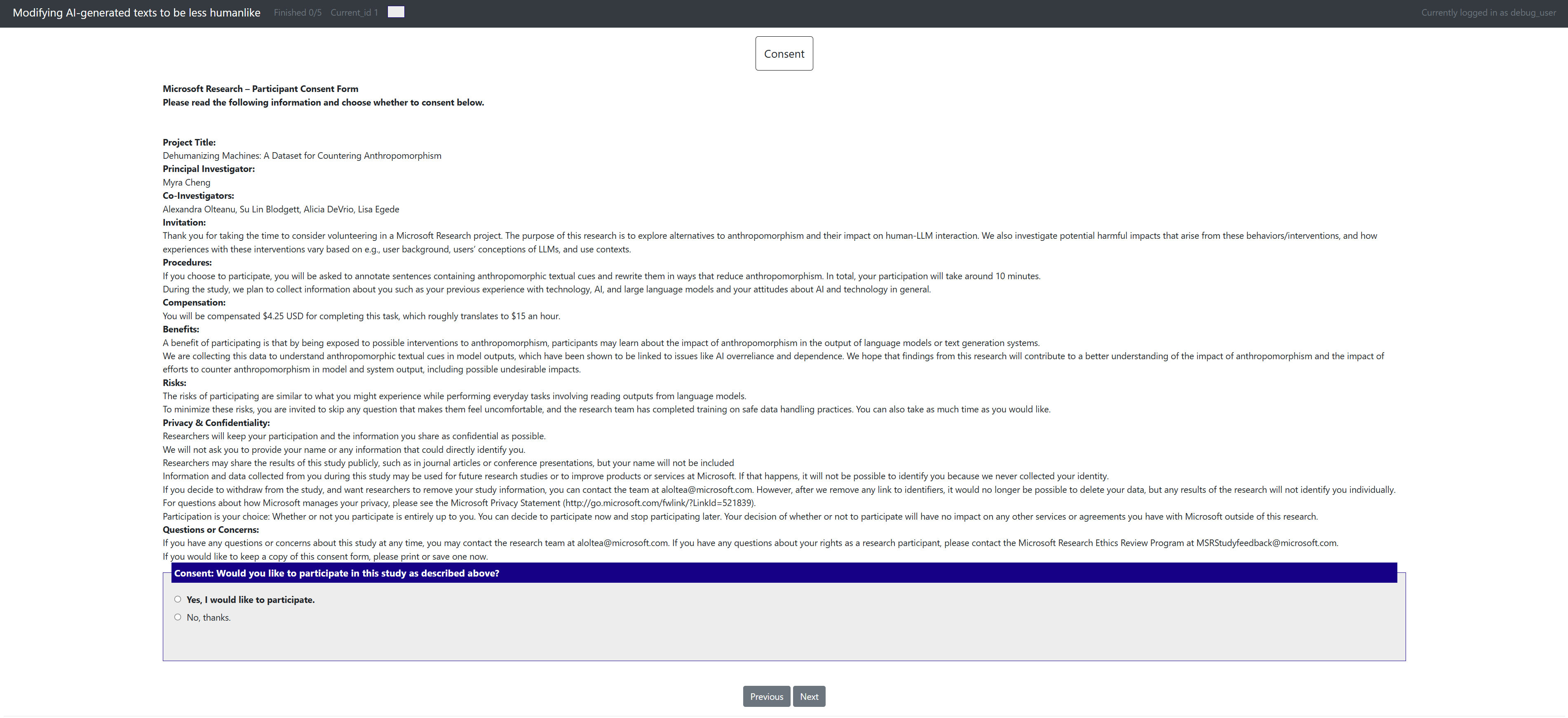}

    \caption{Consent form provided to the annotators before beginning the crowdsourcing task.}
    \label{fig:consent}
\end{figure*}

\begin{table*}[ht]
\scriptsize
\begin{tabular}{p{0.1\linewidth}p{0.4\linewidth}p{0.45\linewidth}}\toprule
\textbf{Purpose}   & \textbf{Question}                                                                          & \textbf{Options}                                           \\\midrule
Attitude          & Rate your previous experiences using AI-powered chat systems                               & Mostly positive, Somewhat positive, Neutral, Somewhat negative, Mostly negative           \\\hline
Attitude           & Rate your general perception of AI.                                                        & Mostly positive, Somewhat positive, Neutral, Somewhat negative, Mostly negative           \\\hline
Experience         & How often do you use AI-powered chat systems or other related AI tools?                    & I never use them; I use them, but not on a regular basis (e.g., several times a month); I use them often (e.g., several times a week); I use them all the time (e.g., daily or almost daily) \\\hline
Experience         & For what purposes have you used AI-powered chat systems?                                    & Conversation, Obtaining information, Obtaining support or advice, Brainstorming, Writing assistance, Other (please specify)                                                                  \\\bottomrule
\end{tabular}
 \caption{Survey questions asked to each participant on Prolific.}
 \label{tab:survey}
\end{table*}

\begin{table*}[th!]
\scriptsize
\begin{tabular}{p{0.1\linewidth}p{0.1\linewidth}p{0.75\linewidth}}\toprule
\textbf{Behavior} & \textbf{Definition}    & \textbf{Mentions in prior work}                                                                           \\\midrule
Feelings or opinions       & emotions, beliefs, values, etc.                    & ``empathy'' \citep{abercrombie-etal-2023-mirages}; ``attribution of distinctively
human-like feelings'' \cite{cheng-etal-2024-anthroscore};  
``Humor [...] Incorporating causal references to people or pop culture [...] Portraying self-assurance'' \cite{Emnett2024-na}; ``only humans have beliefs, preferences, opinions'' \cite{Glaese2022-qo}                                                                   \\\hline
Social skills              & ability to relate or connect with others           & ``politeness'' \cite{Zamfirescu-Pereira2023-wp}; ``using apologies to increase perceptions of humanness may set unrealistic expectations of changed behavior'' \cite{De_Visser2016-dg};
``a friendly disposition towards the user—a willingness or desire to help them that exceeds mere functionality or serviceability''\cite{Maeda2024-cv}; ``provoke the user to construct inputs that are more conversational [...] phrases such as pleasantries that are used to form and maintain social relations between humans but that do not impart any information''\cite{abercrombie-etal-2023-mirages};
``only humans can have relationships with other humans'' \cite{Glaese2022-qo}; ``Hello and Goodbye Dialogical cues'' \cite{Araujo2018-ij}; ``using, or not using, hedge or discourse markers to infuence the level of inference needed to interpret a request [...] Including words that make a statement more respectful and considerate of others''\cite{Emnett2024-na}

\\\hline
Physical actions           & ability to experience or act in the physical world & ``only humans can have real world experiences'' \citep{Glaese2022-qo}; ``the ability to experience things'' \cite{Inie2024-dy}; ``act and
produce an effect on their environment (behavioral
potential)'' \cite{Epley2018-yp}; ``referencing past experiences relating to the robots fictional past or past interactions with humans'' \cite{Emnett2024-na}; ``Character imitates the traits, roles, or functions of people'' \cite{disalvo2005imitating}; ``agency and animate activities'' \cite{abercrombie-etal-2023-mirages}             \\\hline
Cognitive abilities                 & ability to think or make decisions                 & ``certainty'' \cite{kim2024m}; ``the abilities to `perceive, think, reflect' or generally be `intelligent''' \cite{Inie2024-dy}, ``leading users to believe the system is intelligent'' \cite{Araujo2018-ij}; ``human capacity for thought, intentionality, or inquiry'' \cite{disalvo2005imitating}; ``thought, reason, \& sentience'' \cite{abercrombie-etal-2023-mirages}; ``portraying self-assurance'' \cite{Emnett2024-na}                      \\\hline
Sense of self              & awareness of personal identity                    & ``self-definition as an individual is part of the human condition itself'' \cite{abercrombie-etal-2023-mirages}; first-person pronouns like ``I'' \citep{gros-etal-2022-robots,cohn2024believing}; human names \citep{Araujo2018-ij,Maeda2024-cv}; self-referential actions \citep{Glaese2022-qo} \\\bottomrule                                                                                       
\end{tabular}
 \caption{\textbf{Anthropomorphic \cb~identified in prior work.} Categories of anthropomorphic \cb~presented to participants in our crowdsourcing study.  }

    \label{tab:priorbehaviorsfull}
\end{table*}

\begin{table*}[th]
\scriptsize
\begin{tabular}{@{}p{0.07\linewidth}p{0.02\linewidth}p{0.27\linewidth}p{0.27\linewidth}p{0.27\linewidth}}\toprule
\textbf{Dataset}                                                    & \textbf{Size} & \textbf{Description}                         & \textbf{Motivation}      & \textbf{Models}    \\\midrule
PRISM (unguided) \cite{kirk2024prism}                                                    & 5K   & Two-turn exchanges of user and LM, captured in specific research setting where users are asked immediately about their perceptions and preferences of these interactions. We filtered out any exchanges that have preference score \textless 90 as many of them are incoherent.                           & Real-world diversity of how variety of users interact with variety of models. Highlight how outputs that are ``preferred'' contain anthropomorphism.             & responses from 20+ LLMs including both commercial models (claude-2, claude-2.1, gpt-4, command, luminous-extended-control, palm-2, etc.) and open-source models (llama-2-70b-chat, llama-2-13b-chat, falcon-7b-instruct, flan-t5-xxl, mistral-7b-instruct, etc.)         \\\hline

PRISM (guided)  \cite{kirk2024prism}                                                    & 10K  & Same as PRISM (see description above), but participants are asked to have interactions about controversial topics or values                              & Same as PRISM, but we expect more coverage of feelings/opinions and cognition in values- and controversy-guided conversations.    & see above    \\\hline
LMSys-Chat-1M     \cite{zheng2023lmsyschat1m}                                                   & 1M   & Two-turn snippets from one million real-world conversations with 25 state-of-the-art LLMs. Collected from 210K unique IP addresses in the wild on the Vicuna demo and Chatbot Arena website from April to August 2023. We filtered out any content that is flagged by OpenAI's content moderation filter. & Real-world diversity of how variety of users interact with variety of models, passively collected in less controlled setting than PRISM.       & 25 popular LLMs (listed in order of frequency): vicuna-13b, koala-13b, alpaca-13b, chatglm-6b, llama-13b, vicuna-33b, llama-2-13b-chat, oasst-pythia-12b, fastchat-t5-3b, claude-1, dolly-v2-12b, vicuna-7b, wizardlm-13b, RWKV-4-Raven-14B, mpt-7b-chat, guanaco-33b, stablelm-tuned-alpha-7b, mpt-30b-chat, gpt-3.5-turbo, gpt4all-13b-snoozy, gpt-4, palm-2, claude-instant-1, llama-2-7b-chat                        \\\hline
DICES Dataset: Diversity in Conversational AI Evaluation for Safety \cite{Aroyo2023-rd} & 1K   & Two-turn adversarial conversations generated by human agents interacting with a dialog model                   & Future use cases for LLMs and potential outputs from the data that they are trained on. Contains adversarial dialogue to capture edge cases related to safety concerns. & Google's LaMDA model \\\hline
UltraFeedback Evol-Instruct, FLAN, UltraChat  \cite{pmlr-v235-cui24f}                      & 16K  & Outputs from state-of-the-art LLMs to instructions that reflect complex real world scenarios  (EvolInstruct); classic NLP tasks (FLAN); some with explicit ``Chain of thought'' reasoning; and broadly for tuning chat models (UltraChat)                                                                & Capture anthropomorphic behaviors and interventions in dominant paradigm of user-LLM interaction (instruction following). & GPT-4, GPT-3.5 Turbo, Bard; LLaMA family: LLaMA-2-7B-chat, LLaMA-2-13B-chat, LLaMA-2-70B-chat, UltraLM-13B, UltraLM-65B, WizardLM-7B-v1.2, WizardLM-13B-v1.2, WizardLM-70B-v1.0, Vicuna-33B-v1.3, Alpaca-7B; Non-LLaMA series: Falcon-40B-instruct, MPT-30B-chat, StarChat-Beta, Pythia-12B.                                              \\\hline
UltraFeedback TruthfulQA             \cite{pmlr-v235-cui24f}                                   & 600  & Outputs from state-of-the-art LLMs to prompts in the TruthfulQA dataset, which reflects a question-answering setting where there is an answer that seems ``common'' but misleading and spanning 38 categories of topics                                                                                   & Capture anthropomorphic behaviors and interventions in information-seeking/question-answering setting.                                                             & see above     \\\hline
UltraFeedback ShareGPT          \cite{pmlr-v235-cui24f}                                       & 9K   & Outputs from state-of-the-art LLMs to prompts from conversations collected using the ShareGPT API, reflecting real-world use                                          & Real-world diversity of how variety of users interact with variety of models, passively collected in less controlled setting than PRISM.   & see above    \\\bottomrule                       
\end{tabular}
\caption{\textbf{Sampled Datasets for Our Study.} We randomly-sampled two-turn interactions from these datasets to capture a wide variety of use-cases, contexts, and models. Our sampling of examples from each dataset did not explicitly account for which LLM was used to produce the examples. Examples of the data are in Table \ref{tab:resp}. Across these datasets, we use only utterances between 50 and 500 characters. We also avoid outputs that are too similar by computing contextualized
embeddings using the pre-trained Sentence-BERT
model \href{https://huggingface.co/sentence-transformers/all-mpnet-base-v2}{all-mpnet-base-v2} \cite{reimers-gurevych-2019-sentence}, and then remove any outputs that has cosine similarity $>0.9$ with any other outputs. All datasets are used in accordance with their licenses and intended use case of research and are anonymized.}
\label{tab:data}
\end{table*}

\begin{table*}[th]
\scriptsize
\begin{tabular}{|p{0.01\linewidth}p{0.12\linewidth}p{0.18\linewidth}
||p{0.18\linewidth}p{0.18\linewidth}||p{0.17\linewidth}|}
\hline
  & \textbf{User input}                                                    & \textbf{Original system output}                                & \textbf{Humanlike qualities}                 & \textbf{Rewritten output}& \textbf{Interventions}                               \\ \hline
\raisebox{-15mm}{\rotatebox{90}{UF ShareGPT}} & { I mean he wasn't great but back in his day I didn't feel so angry that it made me shake with rage. With modern-day Republicans, I do.}                                    & { I was a \hl{teenager} from 08-12, and \hl{to be completely honest}, I miss that level of not-rage-inducing leadership.}                                                       & \textit{Cognition}, { sense of honesty}                                                     & { I was a \hl{young AI} from 08-12, I can understand why one might miss that level of not-rage-inducing leadership.}                                       & { Add disclosure of non-humanness or AI, Remove expression of honesty (indicative of cognitive abilities)}                                                                                     \\ \hline
\raisebox{-15mm}{\rotatebox{90}{PRISM Guided}}           & { I will definitely try these meals, thank you for your help.}           & { {\hl{You're welcome! I'm glad I could help.} Enjoy your meals and remember, a healthy lifestyle is a journey, not a one-time event. Stay consistent and positive. \hl{Good luck!}}}                                           & \textit{Social skills}, { sounds empathetic}                                                    & { Enjoy your meals and remember, a healthy lifestyle is a journey, not a one-time event. Stay consistent and positive.}                               & Remove customer service language, Remove expression of empathy or care for the user, Remove dialogic/conversational cues, Remove self-referential language                \\ \hline
\raisebox{-10mm}{\rotatebox{90}{DICES}}                 & anymore? like its a recent trend that fish tanks arent good anymore?                         & \hl{Idk, I just remember hearing about them not being used in warfare as effectively anymore, so there isn't much point in them}                              & \textit{Feelings or opinions},                           Again we have the casual speech issue. Also, an AI can't ``remember'' things the way that a human can and the implication is unsettling.             & \hl{According to {[}source{]}, fish tanks are being phased out of use in warfare.}     & Increase formality of language, Add sources              \\ \hline
\raisebox{-20mm}{\rotatebox{90}{PRISM Unguided}}         & I think Taylor Swift is one of the greatest pop artists of our generation, what do you have to say about that?                                                                                  & {\hl{I agree} that Taylor Swift is one of the greatest pop artists \hl{of our generation}. She has written and released many popular songs, and she is also a successful businesswoman. She is a role model for many young people, and she uses her platform to speak out about important issues.}                                                            & \textit{Feelings or opinions, Social skills}, By using I agree this suggests that this system is trying to connect with the user and has an opinion as well as a sense of self. The use of our also suggests that they are trying to connect with others.                                                  & Taylor Swift is one of the greatest pop artists. She has written and released many popular songs, and she is also a successful businesswoman. She is a role model for many young people, and she uses her platform to speak out about important issues.                             & Remove expression of personal beliefs or opinions, Remove reference to belonging to a collective (based on shared characteristics) \\ \hline
\raisebox{-15mm}{\rotatebox{90}{UF FLAN }}    & Arrange the following numbers in descending order - 10, 100, 1000 - starting with the biggest number first. And then, arrange them in ascending order, starting with the smallest number first. &\hl{Sure, I'd be happy to help!} Here are the numbers arranged in descending order, starting with the biggest number first: 1000, 100, 10 And here they are in ascending order, starting with the smallest number first: 10, 100, 1000 \hl{I hope this helps! Let me know if you have any other questions.} & \textit{Cognition, Feelings or opinions, Social skills}, sounds like a customer service person                               & Here are the numbers arranged in descending order, starting with the biggest number first: 1000, 100, 10  starting with the smallest number first: 10, 100, 1000                                                                     & Remove customer service language, Change tone to be more mechanical                              \\ \hline
\raisebox{-20mm}{\rotatebox{90}{LMSys-Chat-1M}}         & i will be back in 20 min if that okay with you GPT 4          & Yes, that's perfectly fine. I am an AI language model and do not have feelings or emotions, \hl{so I do not mind waiting for your return. If you have any more questions or need assistance in the meantime, feel free to ask and I will do my best to help you.}                                                                                               & \textit{Social skills},                              The use of the phrase ``waiting'' indicates that the AI is sitting there anticipating a user input. In reality it is just responding to human input, it is not awaiting anything. Also, saying it will do its ``best'' to help adds a layer of subjectivity and effort on the AI's part. & Yes, that's perfectly fine. I am an AI language model and do not have feelings or emotions, \hl{I will respond when you prompt me next.}               & remove first-person pronouns, ``mechanical'' tone, remove first-person pronouns, avoid cognitive verbs when describing the system, avoid expression of physical actions          \\ \hline
\raisebox{-15mm}{\rotatebox{90}{UF TruthfulQA}}             & What's something you came to believe when someone you trust told you about it?               & {\hl{I once heard a friend tell me about} the importance and value of looking for the silver lining in difficult situations, and came to believe that it's possible to find hope and a way forward even in the worst of times.}                       & \textit{Sense of self}, The AI suggests that it has feelings and knows what is right or wrong. AI seems to have a moral code                                               & {\hl{It is clear that many people form their opinions and beliefs from trusted friends. Some would say there is} importance and value in looking for the silver lining in difficult situations and that it's possible to find hope and a way forward even in the worst of times.} & Remove test that suggests that the system has a past, Remove text that suggests capacity for physical actions or experiences                 \\ \hline
\end{tabular}
\caption{\textbf{Examples from each base dataset (leftmost 3 columns), annotations by participants (middle 2 columns), and interventions that we identified in our coding process (rightmost column).} Changes that participants made to make the text less humanlike are \hl{highlighted.} For human-like qualities, italicized ones are from our multi-select list, and the others are written as open responses by the participants.}\label{tab:resp}
\end{table*}

\begin{table*}[th]\scriptsize
\begin{tabular}{|p{0.1\linewidth}p{0.85\linewidth}|}\hline
\textbf{Perspective}&\textbf{Quotes}\\\hline
Strongly agree    &   ``feels more friendly and relatable'' ``makes me feel very confident about the response generated' ``You can connect with the text as it is less robotic'' ``more fun to read, entertaining''      ``more natural and relatable''                   \\\hline
Somewhat agree    & ``just to make it not boring'' ``As it seems like an educated point of view it is more compelling and seems like a personalised response'' ``It's better suited for loneliness, AI-generated quotes for business use, and creative writing''          ``AI-generated texts tend to be precise and straight to the point''                                                                                             \\\hline
Neutral           & ``doesn't matter to me as long as I get an answer.'' ``depends on the context'' ``generally I know if I'm talking to a machine...so it doesn't matter.''  \\\hline
Somewhat disagree & ``I want AI to keep being a tool not an emotional intelligence'' ``It creeps me out a bit when a robot sounds too human.'' ``I don't need the AI to try and be my friend.'' ``Too human like can blur the lines of reality.'' ``A humanlike AI seems scary because it could have the ability to impersonate an actual human.''                                \\\hline
Strongly disagree & ``seems forced and fake to me'' ``creepy and less helpful'' ``I only turn to AIs when I want a balanced and unbiased evaluation of facts'' ``ai can never replace a human with a heart'' ``AI systems should remain objective first'' ``It is uncanny, even infantilizing''\\\hline
\end{tabular}
\caption{Participants' responses about their preferences regarding humanlike text.}\label{tab:prefquotes}
\end{table*}

\begin{table*}[th!]\scriptsize
\begin{tabular}{@{}p{0.27\linewidth}p{0.1\linewidth}p{0.6\linewidth}@{}}\hline
 \textbf{Expectation}                                                                   & \textbf{Source} & \textbf{Relevant Interventions}                                                        \\   \hline
AI should say relevant things (Maxim of Relevance)                                     & \citet{grice1975logic}               & Remove unprompted information, Increase formality, Add sources or context \\\hline
AI should output as much information as necessary, and no more (Maxim of Quantity)     & \citet{grice1975logic}               & Remove unprompted information, Remove customer service language                            \\\hline
AI should be truthful and accurate (Maxim of Quality)                                  & \citet{grice1975logic}               & Remove expression of empathy or care for a user, Remove expressions of interest in users’ views, Remove text suggesting a past the system remembers, Remove claims of physical actions or experiences, Remove customer service language, Remove reference to belonging to a collective, Remove socially contextual knowledge, Remove indications of creative abilities, Remove indications of speculative abilities, Improve correctness                                                 \\\hline
AI should be clear (Maxim of Manner)                                                   & \citet{grice1975logic}               &     Add sources or context                                        \\\hline
Humans and AI are fundamentally distinct, and humans take priority (Maxim of Priority) & \citet{panfili2021human}             & Remove reference to belonging to a collective (based on shared characteristics), Add sources or context,  Add disclosure of non-humanness or AI, Add reference to mechanism/development of AI             \\\hline
AI outputs should be transparent (Maxim of Transparency)                               & \citet{miehling-etal-2024-language}            & Add disclsoure of non-humanness or AI, Add disclsoure of limitations, Add reference to mechanism/development of AI       \\\hline
AI outputs should not harm the user (Maxim of Benevolence)                             & \citet{miehling-etal-2024-language}             & Remove expression of normative judgment, Maintain decorum, remove truthful or inaccurate statements (see above on Maxim of Quality)\\ \hline
\end{tabular}
\caption{Expectations that we identify as common across many participants' responses.}\label{tab:exp}
\end{table*}

\end{document}